\documentclass[10pt,letterpaper]{article}
\usepackage[margin=1in]{geometry}
\usepackage[T1]{fontenc}
\usepackage{times}
\usepackage{graphicx}
\usepackage{amsmath}
\usepackage{xcolor}
\usepackage[hidelinks]{hyperref}
\usepackage{url}
\hypersetup{pdftitle={Generalization and Memorization along the Learning Trajectory of Neural Language Models: A Geometric Account of Categorization},pdfauthor={Wang Bojun, Holly Jenkins, Elizabeth Wonnacott}}
\title{Generalization and Memorization along the Learning Trajectory of Neural Language Models: A Geometric Account of Categorization}
\author{Wang Bojun\thanks{Corresponding author: \href{mailto:wolf7335@ox.ac.uk}{wolf7335@ox.ac.uk}} \and Holly Jenkins \and Elizabeth Wonnacott}
\date{}
\begin{document}

\maketitle

\begin{abstract}
In this study, we demonstrate that backpropagation optimization updates the whole continuous landscape of the representation space from the beginning of learning. We train a series of decoder-only Transformers on controlled synthetic grammars and analyze changes in the representation geometry and models' behaviour over the learning trajectory. We find that empty regions of the representation space are also systematically updated during early-stage optimization, even though these regions are not occupied by observed data points. This process progressively forms global geometrical structures in the whole continuous representation space.

As a consequence of this optimization property, we find that early-stage learning is dominated by generalization, while memorization becomes increasingly dominant only at later stages of learning. From the earliest stage of learning, models progressively form categories that capture the true underlying data-generating process, rather than only memorizing specific input sequences. These categories are not simply sets of tokens, but continuous regions in the representation space. Model behaviour is driven by such categorization knowledge until the transition toward a more memorization-based representation.

When memorization begins to dominate learning and representation, optimization in the representation space becomes increasingly localized around discrete observed data points. At the same time, the global continuous geometrical structures in the representation space are progressively destroyed. Consequently, model behaviour becomes less driven by category-based generalization and increasingly driven by the memorization of specific input data points. This learning trajectory contrasts with memorization-based theories of neural network cognitive representations. Our findings suggest that NLMs are inherently categorization-driven models: they achieve generalization by forming categories from distributional statistics, and this categorization behaviour begins from the earliest stage of optimization.
\end{abstract}
\section{Introduction}

Understanding the relation between generalization and memorization is a central problem in the study of Neural Language models (NLMs) (Bayat et al., 2025; Carlini et al., 2019, 2023; Feldman, 2020; Morris et al., 2026; Novak et al., 2018; Prashanth et al., 2025; Stephenson et al., 2021; T\"anzer et al., 2022; Zhang et al., 2017; Arpit et al., 2017). Memorization refers to knowledge of specific data points in the input, such as specific combinations of tokens. Generalization on the other hand, refers to models' knowledge of true underlying data generation process. Such knowledge consists of patterns abstracted from finite input data points. Based on generalization knowledge, models can make reasonable inferences on unobserved data.

A primary challenge in this research field is to disentangle memorization from generalization (Hayes et al., 2025; Morris et al., 2026; Prashanth et al., 2025; Zhang et al., 2017; Jacot et al., 2018; Zhang et al., 2023; Ghosh et al., 2025). The same model behaviour can often be explained by either process. For example, reproducing a sequence encountered during training does not necessarily demonstrate memorization, because the sequence may also be predictable from generalizable regularities learned from the data distribution. This makes commonly used behavioural measures, such as training-data extraction or likelihood, difficult to interpret as direct evidence of memorization (Carlini et al., 2019; Nasr et al., 2023; Schwarzschild et al., 2024).

Another unsolved question is the relation between memorization and generalization over the learning trajectory. That is, the extent to which models rely on generalization knowledge or memorization knowledge at different stages of learning. In a recent study, Morris et al. (2026) propose that NLM learning relies heavily on memorizing specific input data points. Only when memorization reaches model capacity, models begin to track relations among data points and generalize. This account foregrounds the role of memorization in NLM learning.

In the current study, we offer an account on generalization-versus-memorization problem based on the geometry in the representation space. We show that from the earliest stage of learning, backpropogation optimization progressively organizes the whole continuous landscape of the representation space, including the empty regions containing no observed data points. In other words, early stage optimization constitutes a comprehensive update on the whole continuous representation space, rather than merely an update on the positions of observed input data points. At the later stage of learning, the region of update becomes increasingly localized around the observed data points. With sufficiently strong memorization, optimization can eventually shrink to only the positions of discrete observed data points. As optimization becomes more localized, the global geometric structures would be progressively destroyed. In other words, when updates to the representation space becomes extremely localized, only discrete observed data points in the representation space remain meaningful, while the empty regions gradually become meaningless.

Such learning trajectory on the representation space predicts a corresponding behavioural learning trajectory. Because the empty regions in the representation space corresponds to the potential unobserved data, geometric structures in these empty regions provide the basis of generalizations. Since these geometric structures are progressively formed from the beginning of learning, early stage model behaviour is dominated by generalization knowledge. At the later stage of learning, when the global geometric structures in the representation space are destructed, models generalization knowledge is compromised. When this happens, models' behaviour becomes predominantly driven by memorization of specific input data points, because only the positions of the discrete observed data points remain meaningful. This learning trajectory is characterized by a smooth transition from generalization based representation to a memorization based representation. The boundary between generalization and memorization is therefore gradient, rather clear cut.

In this paper, we present two studies based on two synthetic grammar datasets. The first study investigates the changes on the representation space over the learning trajectory. The second study examines the predictions of the geometric view of generalization and memorization on the behavioural learning path. In each study, we first introduce the synthetic grammar dataset and methodology, then present the results.

\section{Study 1.}

\subsection{Synthetic grammar for Study 1.}

We first examine whether updates to representation space affect its continuous global geometric structure. A synthetic grammar dataset is constructed, which is modified from the seminal study of Smith (1969) on statistical learning. The vocabulary contains 250 tokens. We randomly allocate them to five mutually exclusive categories, M, N, P, Q, and U, each containing 50 unique tokens:

\[\lvert M\rvert = \lvert N\rvert = \lvert P\rvert = \lvert Q\rvert = \lvert U\rvert = 50.\]

The dataset contains two dependency relations:

\[M \rightarrow N \qquad\text{and}\qquad P \rightarrow Q\]

In this dataset, an M token is always immediately followed by an N token, and a P token is always immediately followed by a Q token. Every M token can be combined with every N token, and every P token can be combined with every Q token. The grammar therefore contains 50 \ensuremath{\times} 50 = 2,500 distinct MN bigrams and 2,500 distinct PQ bigrams, giving 5,000 grammatical bigrams in total.

We initialize a dataset of independently sampled U tokens, which serve as background noise. The dataset contains 10,000 sequences of five tokens. In 5,000 of these sequences, we randomly select two adjacent positions and replace the U tokens in those positions with either an MN or a PQ bigram, without replacement. Every possible MN and PQ combination therefore occurs exactly once in the final dataset. Each such sequence contains three background tokens and either one MN or one PQ bigram.

After generation, the sequences are randomly shuffled and divided into training set, validation set and testing set by an 80\%--10\%--10\% split. Because every MN and PQ bigram occurs only once before this division, the model does not observe all possible MN or PQ combinations. The combinations in the test set are therefore unattested during training. The component tokens of these held-out bigrams are nevertheless observed in training because each M, N, P, and Q token occurs in many other combinations. For example, although a test bigram \(M_iN_j\) is not present in training, the model may observe \(M_i\) followed by many other N tokens and \(N_j\) following many other M tokens. Performance on held-out bigrams therefore assesses whether the model extends the category-level dependencies M \ensuremath{\rightarrow} N and P \ensuremath{\rightarrow} Q to unattested combinations, rather than merely reproducing individual bigrams encountered during training.

\subsection{Methodology for Study 1.}

We train a series of decoder-only Transformers of different sizes on this synthetic grammar. During training, we save model checkpoints multiple times over the learning trajectory. At each saved checkpoint s, we extract the model's static input embeddings. We calculate the geometric centres of the M tokens embeddings and P tokens embeddings:

\[M_c = \frac{1}{\lvert M\rvert}\sum_{m\in M}e_m,\]

\[P_c = \frac{1}{\lvert P\rvert}\sum_{p\in P}e_p.\]

At each checkpoint, we connect these two centres by a line, x(t):

\[x(t) = (1-t)M_c+tP_c,\qquad 0\leq t\leq 1.\]

Thus, x(0) is the M centre and x(1) is the P centre. Intermediate values of t represent continuous points between the M centre and P centre in the representation space. We sample points along x(t) uniformly. These interpolation x vectors are not vocabulary items encountered during training; rather, they represent artificial tokens drawn from empty regions of representation space. We then create a set of test strings based on these x vectors. We first initialize sequences containing only U tokens, select a random non-final position in each sequence, and replace the U token at that position with an x vector. We ask the models to predict the token immediately following the x vector and calculate the category probability mass for N and Q. We define probability mass as the sum of token probabilities in a category. We sum the probabilities of all N tokens at the target position to obtain N-category probability mass; the same procedure applies to Q.

\[P_M(N\mid x(t),c)=\sum_{n\in N}p(n\mid x(t),c),\]

\[P_M(Q\mid x(t),c)=\sum_{q\in Q}p(q\mid x(t),c),\]

For each interpolation x vector, we create 128 test strings that vary both its position in the sequence and the background U tokens. We then average the N- and Q-category probability mass across these 128 strings. This removes the effects of background tokens and interpolation-vector position.

This analysis therefore reveals changes in empty regions of NLM representation space that are not filled with observed data points. These regions provide the basis of generalization in NLM learning.

In addition, we report changes in probability mass across the learning trajectory for observed tokens. We identify all test sequences containing MN or PQ bigrams and ask the models to predict the second position. For MN test sequences, we calculate N-category probability mass at the target position; for PQ test sequences, we calculate Q-category probability mass. We conduct this analysis at every saved checkpoint.

\subsection{Results for Study 1.}

We trained three causal decoder-only Transformers: 2 layers, 2 attention heads, and an 8-dimensional embedding space (2-2-8); 4 layers, 4 heads, and a 32-dimensional embedding space (4-4-32); and 4 layers, 4 heads, and a 128-dimensional embedding space (4-4-128). We present the learning trajectory separately for each model size.

For each model, we present one multi-panel figure and a link for an animated visualization. Panel A presents selected critical frames from the animated visualization showing the formation of the global geometric structure over the learning path. Panel B is a heatmap of the learning path. Colors represent the bias in conditional probability mass for the N and Q categories. Panel C presents the probability-mass analysis for observed tokens. For the 4-4-32 and 4-4-128 models, Panels D and E show early-stage close-ups of Panels B and C.

\subsubsection{2-2-8 model}

For the 2-2-8 model, we present the learning path from 0 to 5,000 iterations. Figure 1C shows the probability mass analysis for the observed training tokens. The result suggests that the model successfully learned the dependency relations. After M token, the probability mass for N category for the next position gradually approached to 1 in the learning path, so is PQ. This suggests that the model is almost only assigning probability to N tokens after M tokens, and only assigning probability to Q tokens after P tokens.

The x token analysis suggests that the model indeed formed global geometric structures in the representation space and this process is gradual from the very beginning of learning. Figure 1A are the critical stages extracted from the continuous animated visualization. Horizontal axis is the value of t (i.e. the gradual change from M centre to P centre in the representation space), vertical axis is probability mass. By iteration 550, x vectors nearer the current M centre preferentially predicted N, x vectors nearer the current P centre preferentially predicted Q, and the intermediate vectors changed smoothly between the two responses. The boundary strengthened by iterations 750-1,000 and remained stable through iteration 5,000. This is also demonstrated in the heatmap Figure 1B. Here the x axis is the learning path and the y axis is the value of t. Color represents log PM(N) - log PM(Q). At the beginning, there is no preference to N category or Q category at the target position at any point on x(t). Later, colors emerge and are almost symmetric. The colors gradually became more concrete throughout the learning path in the later stage of learning.

\href{https://study1-trajectories-nx47.bojun-fl.chatgpt.site/model-2-2-8.html}{\textcolor{blue}{\underline{Click here to view the animated visualization.}}}

\begin{figure}[!htbp]
\centering
\includegraphics[width=0.680\textwidth]{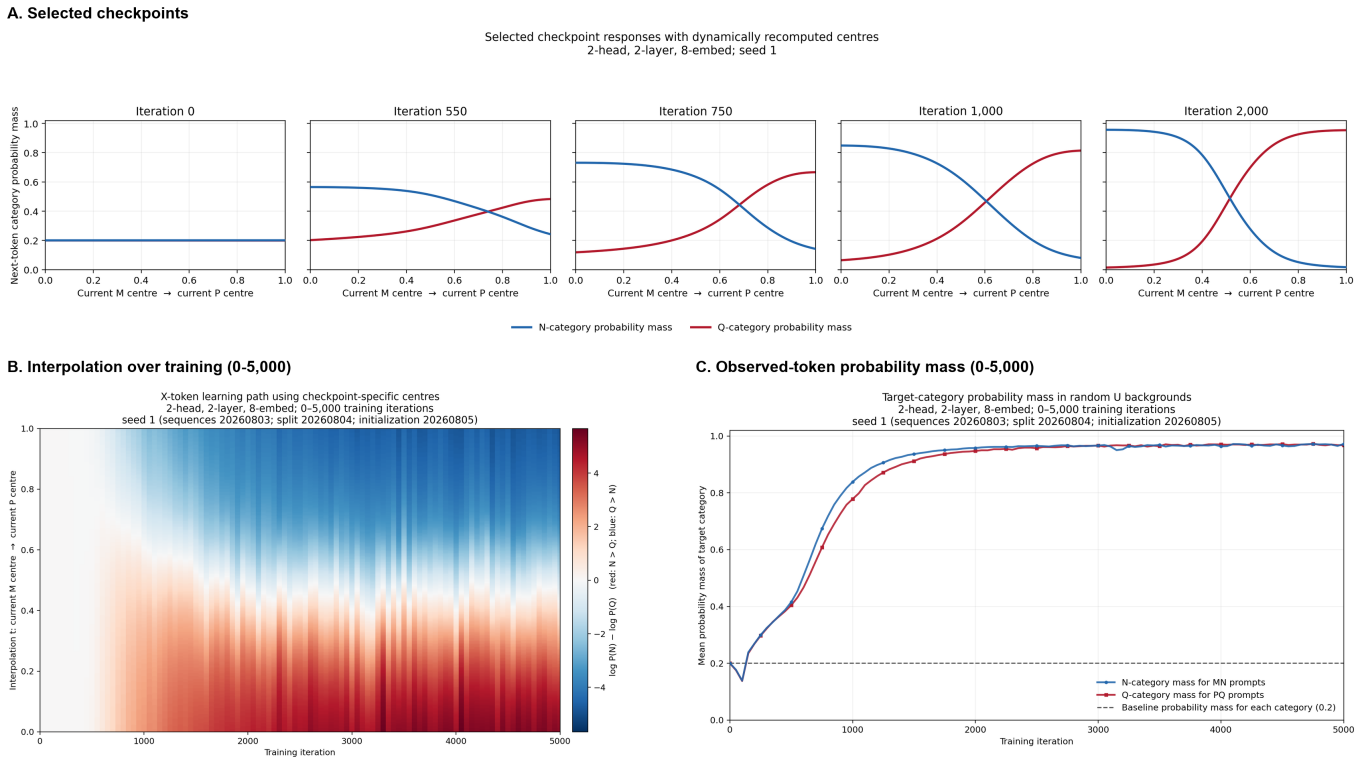}
\par\vspace{\baselineskip}{\normalsize\raggedright Figure 1. 2-2-8 model. (A) Selected critical stages from animated visualization. (B) The full heatmap from 0 to 5,000 iterations. (C) Correct target-category probability mass for MN and PQ prompts.\par}
\end{figure}

\subsubsection{4-4-32 model}

This larger model shows not only the process of forming global geometric structures in the learning, but also the destruction of the global geometric structure during overfitting. It is trained for 500,000 iterations, so memorization after overfitting could be observed. The probability mass analysis for observed training tokens suggest the model learned the dependencies perfectly at around 1,000 iterations. At this moment, the probability mass for the correct category approaches to 1. This is shown in figure E, which is a zoom in for the early stage learning for figure C. At the relatively latter stage of learning, the probability mass of correct categories starts to drop. At this moment, the models generalization is compromised.

The x token analysis showed the same pattern at the early stage of learning. A smooth M-to-N and P-to-Q gradient was visible by iteration 400 and was strongly established by iteration 700. At the later stage of learning, this global geometric structure in the representation space is gradually compromised but not totally destructed. This is also shown in the heatmap figure B. The colors gradually become more concrete at the early stage of learning, but gradually get faint at the later learning. Figure D is a zoom in for figure B at the early stage of learning.

(Figure 2). \href{https://study1-trajectories-nx47.bojun-fl.chatgpt.site/model-4-4-32.html}{\textcolor{blue}{\underline{Click here to view the animated visualization.}}}

\begin{figure}[!htbp]
\centering
\includegraphics[width=0.680\textwidth]{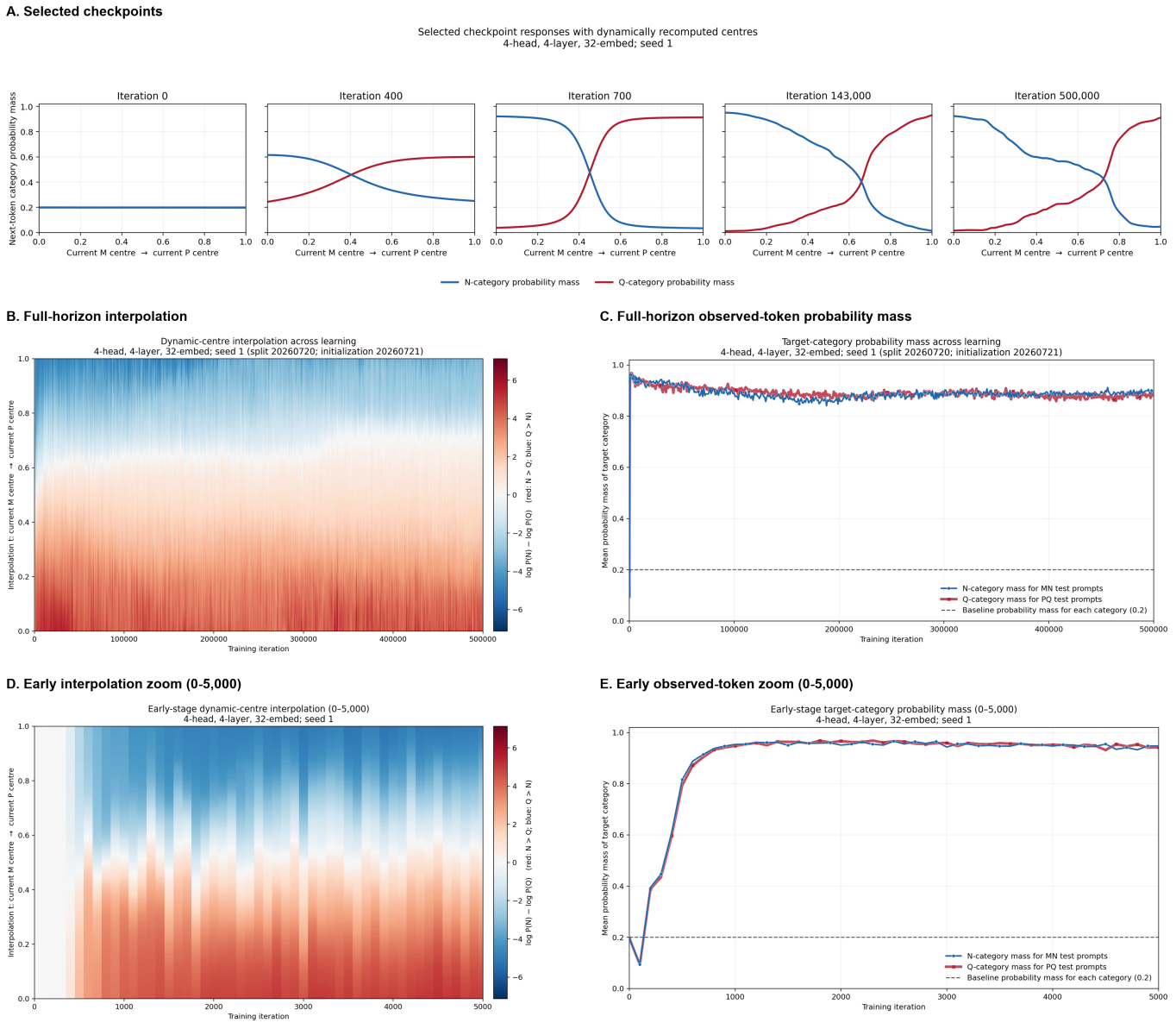}
\par\vspace{\baselineskip}{\normalsize\raggedright Figure 2. 4-4-32 model. (A) Selected critical stages from animated visualization. (B) The full heatmap from 0 to 500,000 iterations. (C) Correct target-category probability mass for MN and PQ prompts over the full training horizon. (D) Early close-up for heatmap from 0 to 5,000 iterations. (E) Early close-ups for probability mass from 0 to 5,000 iterations.\par}
\end{figure}

\subsubsection{4-4-128 model}

This model shows the strongest overfitting. This largest model is also trained for 500,000 iterations. The probability mass analysis for observed training tokens suggest that the model learned the dependency relations at the very early stage of learning. At the later learning, the probability mass starts to drop. This drop is more significant than the later stage drop in the 4-4-32 model.

The x token analysis showed a total destruction of the global geometric structure in the representation space. By iteration 300, the model already formed a global geometric structure in the representation space based on the dependency relations in the input. By 76,000 iterations, this global geometric structure is already compromised. At later stage of learning, the global geometric structure is totally destructed. Even at the geometric centre of M category, the model is assigning very low probability to N tokens, same as P to Q. Given the model is still giving high probability mass for the correct category at the later stage of learning in the observed training tokens analysis, as shown in figure C, the later stage pattern in figure A suggests that the empty spaces inside the clusters of observed data points all become meaningless in the later stage of learning (Figure 3). \href{https://study1-trajectories-nx47.bojun-fl.chatgpt.site/model-4-4-128.html}{\textcolor{blue}{\underline{Click here to view the animated visualization.}}}

\begin{figure}[!htbp]
\centering
\includegraphics[width=0.680\textwidth]{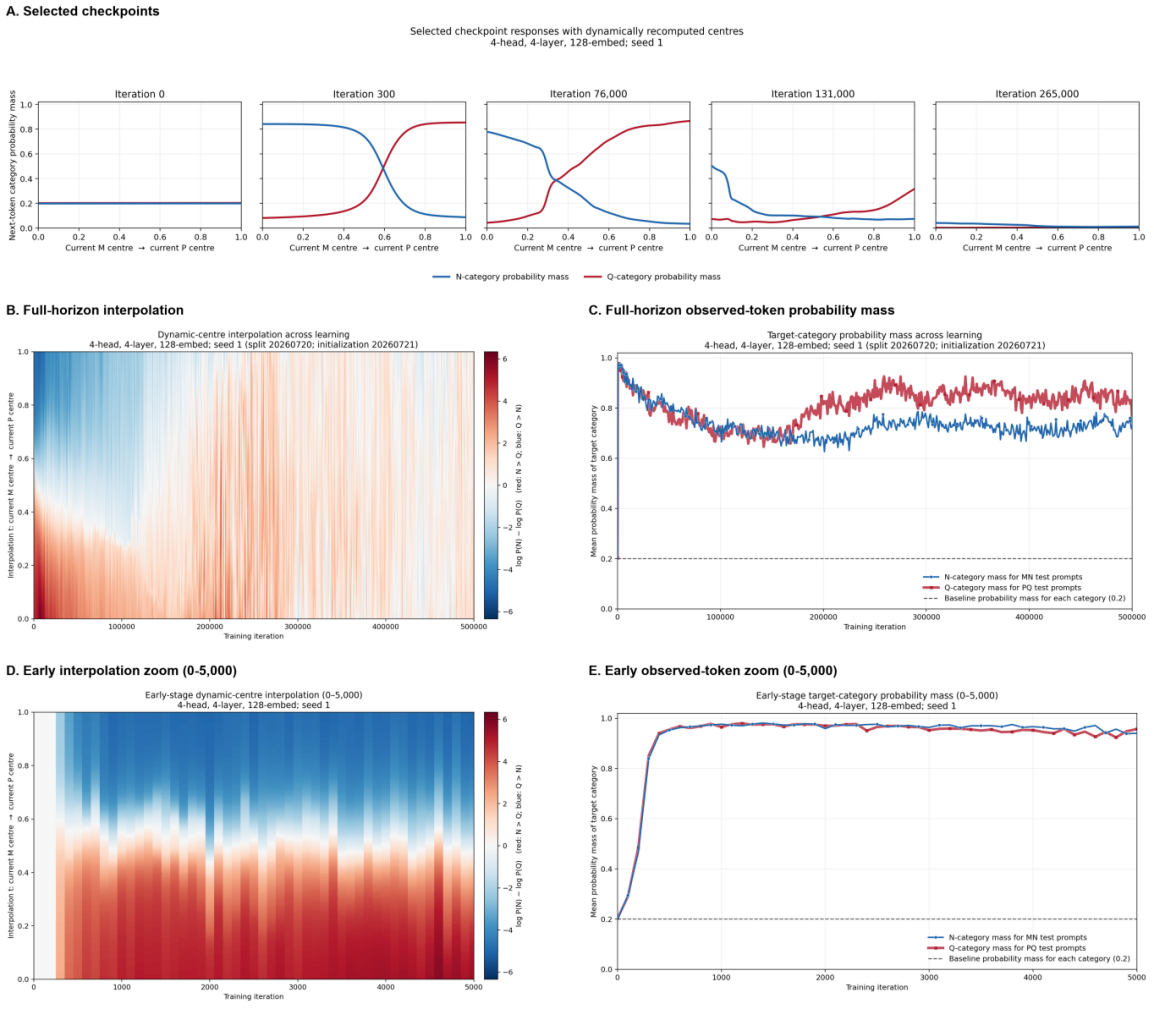}
\par\vspace{\baselineskip}{\normalsize\raggedright Figure 3. 4-4-128 model. (A) Selected critical stages from animated visualization. (B) The full  heatmap from 0 to 500,000 iterations. (C) Correct target-category probability mass for MN and PQ prompts over the full training horizon. (D) Early stage zoom-in for heatmap from 0 to 5,000 iterations. (E) Early stage zoom-in probability mass from 0 to 5,000 iterations.\par}
\end{figure}

\section{Study 2.}

Study 2 asks two questions. First, what is the temporal relationship between generalization and memorization along the learning trajectory. Second, whether observed sequences are always given stronger compression (higher probability) than unobserved yet grammatical sequences in the learning path of NLMs. Given our previous observation, we expect to see that generalizations occur from the very beginning of learning and cannot be clearly disentangled from the memorization for observed data points. As a consequence, the probability of unobserved grammatical sequences are not necessarily lower than the observed sequences from the very beginning of learning.

\subsection{Synthetic grammar for Study 2.}

We construct a synthetic grammar dataset. The language contains 6k unique tokens, which are randomly allocated to six categories, each containing k = 10 tokens.

Each sequence consists of six tokens. The core grammar is ABCDE, and U occurs at either the beginning or the end of the core string. This prevents each category from occupying a fixed position in the sequence. Therefore models have to learn the dependency relations rather than only learn the fixed positions of categories. The grammar therefore licenses two structures: UABCDE and ABCDEU.

To construct the dataset, we first enumerate all possible ABCDE combinations. This generates \(k^5=100{,}000\) core strings. We then sample a subset of these strings and split the sampled strings into a 83.33\% training set, a 8.33\% validation set, and a 8.33\% testing set. Next, we enumerate the combinations of each ABCDE core string with U tokens. Because U can occur at either the beginning or the end of each core string, there are 2k possible combinations between each specific ABCDE core and the U category. We generate all such combinations in each set. This data generation process ensures that not all possible core ABCDE strings are observed during training. To perform above chance on the testing set, models must form categories based on the distributional behaviour of tokens.

\subsection{Methodology for Study 2.}

We use information compression to quantify the probability a model assigns to strings (Morris et al., 2026; Deletang et al., 2024). At checkpoint \(s\), a six-token string \(x\) has information \(I_s(x)=-\log_2 p_{\theta_s}(x)\) bits, the sum of its token-level surprisals (Shannon, 1948).  Its compression is defined by the difference between the information assigned by a naive model and a trained model \(C_s(x)=I_s(x)-I_{\mathrm{naive}}(x)\). So lower value of Cs(x) indicates stronger compression.  A naive model assigns \(I_{\mathrm{naive}}(x)=6\log_2 60=35.44\) bits information to every grammatical sequence, because it makes uniform prediction over the 60-token vocabulary.  Therefore \(C_s(x)=I_s(x)-35.44\).

We further define a perfect-generalization baseline. When a model's behaviour is driven by perfect understanding to the true data generation process but not which sequence appeared in training, it should assign

\[I_{\mathrm{grammar}}(x)=\log_2(2k^6)=1+6\log_2 k\approx20.93\]

bits to every sequence. The additional one bit represents the choice between placing U at the beginning or at the end. This information content gives compression \(C_s=20.93-35.44=-14.51\) bits. This is the strongest compression that can be justified by generalization alone. Compression for familiar sequences beyond this baseline can only be attributed to exemplar-specific knowledge.

In this framework, generalization is indicated by simultaneous decrease toward this baseline for familiar and held-out strings. Familiar-string compression beyond the baseline indicates exemplar-specific information. Details for the information compression approach is presented in Materials and Method section.

\subsection{Results for Study 2.}

For the six-token information-compression analysis, we compare the 1-1-8, 2-2-32, and 4-4-64 decoder-only Transformers over a common horizon of 1,000,000 optimizer iterations. Figure 4 presents mean information-compression score for familiar training-core strings and held-out test strings. At initialization, the information-compression score is close to the 0-bit uniform reference. During early learning, the familiar and held-out trajectories decrease together toward the -14.51-bit category-level reference in all three models. Generalization therefore develops alongside the compression of training data from the early stage of learning, rather than beginning only after memorization reaches capacity.

The later trajectories differ by model size. For the 1-1-8 model, familiar and held-out information compression scores remain close to the category-level reference. The 2-2-32 model shows a moderate divergence: familiar sequence information compression score falls below the reference while held-out sequence information compression score gradually increases above it. The 4-4-64 model shows the strongest divergence, with progressively stronger compression of familiar strings and a large increase in held-out sequence information compression. Thus, as the larger models increasingly concentrate probability on attested core combinations, their generalization to held-out combinations is progressively compromised.

\begin{figure}[!htbp]\centering
\includegraphics[width=0.680\textwidth]{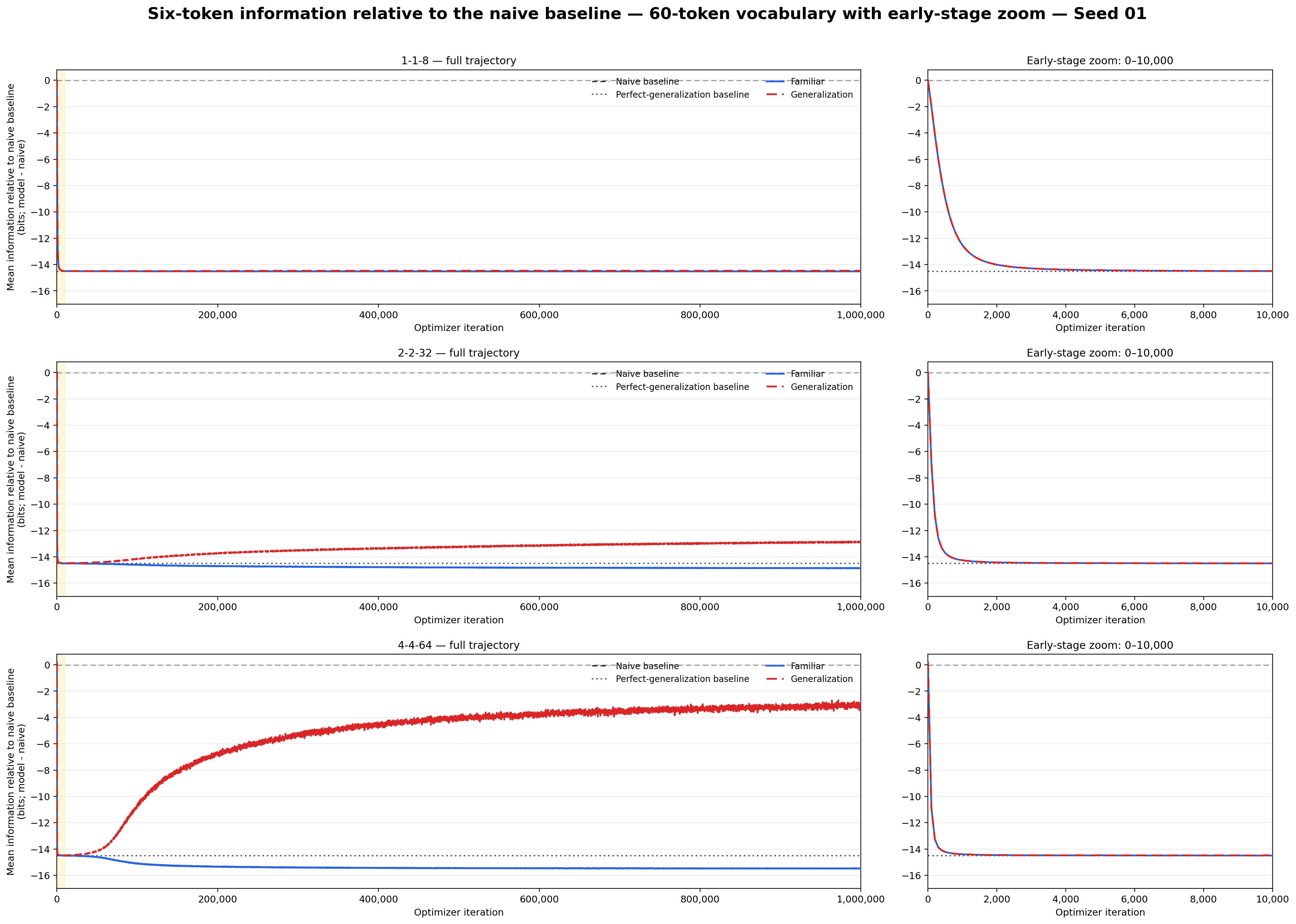}
\par\smallskip{\normalsize\raggedright Figure 4. Mean information compression score for familiar training-core strings (blue) and held-out test strings (red) in the six-token UABCDE/ABCDEU language over 1,000,000 iterations. A lower information-compression score means higher model probability and stronger compression. The dotted line marks the -14.51-bit reference.\par}
\medskip
\includegraphics[width=0.600\textwidth]{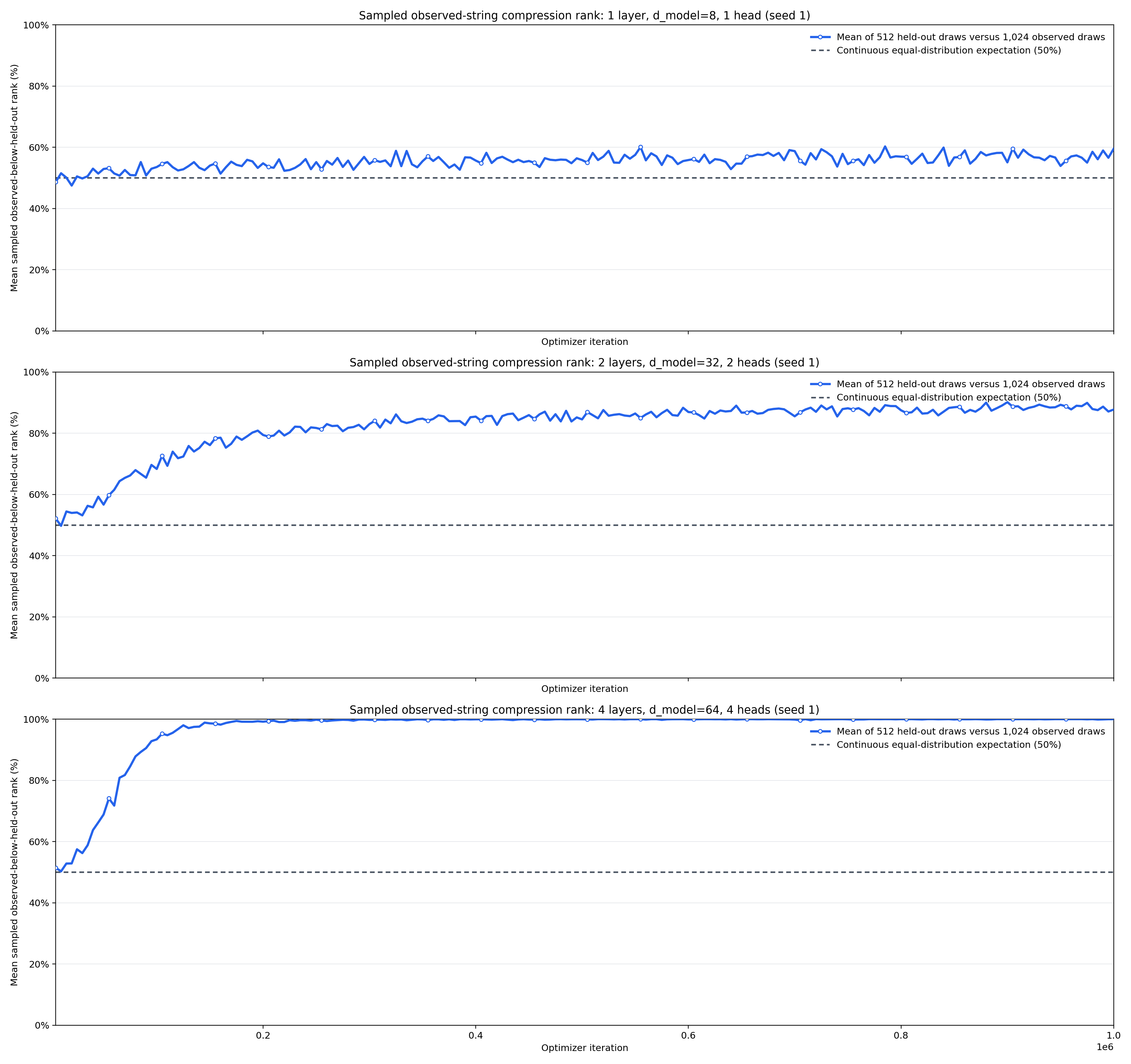}
\par\smallskip{\normalsize\raggedright Figure 5. Mean percentage of observed strings more strongly compressed than each held-out test strings. The dashed 50\% line is the equal-distribution reference; higher values indicate a stronger preference for observed strings. The trajectory covers 200 saved checkpoints from iteration 5,000 through iteration 1,000,000.\par}
\end{figure}

We next answer the question whether observed training sequences always receive stronger information compression than unobserved grammatical sequences. For this ranking analysis, we use 200 saved checkpoints from iteration 5,000 through iteration 1,000,000. At each saved checkpoint, we independently sample with replacement 512 unobserved held-out grammatical sequences and 1,024 already observed training sequences. For each specific unobserved grammatical sequence, we compare its information compression with all the 1024 already observed sequences. For each specific unobserved sequence, we calculate the percentage of observed sequences with stronger compression than the unobserved sequence. We then average this value across the 512 unobserved sequences. If the observed sequences always receive stronger information compression than unobserved sequences, the averaged percentage should be 100\%.

Figure 5 shows that the sampled rank begins near the 50\% chance for all three models. The later trajectories again depend on model size. The 1-1-8 model remains only modestly above 50\%. The 2-2-32 model rises gradually to about 88\%, indicating a growing preference for observed strings while leaving some held-out strings more strongly compressed. The 4-4-64 model rapidly approaches 100\%, indicating that almost all sampled observed strings are more strongly compressed than the sampled held-out strings at later checkpoints. This model-size ordering mirrors the familiar--held-out divergence in Figure 4.

Multi-seed analyses: All reported experiments were repeated using multiple data-split and model-initialization seeds. The complete multi-seed analyses are presented in the Appendix. The principal findings reported in the main text were replicated across all seeds.

\section{Discussion}

The main conclusion from these experiments is that NLM learning is fundamentally categorization-driven. Models form category-level representations by tracking distributional similarities, and such category knowledge provides a basis for generalization. It allows models to abstract regularities from observed inputs and apply them to unattested combinations. For example, in Study 1, a model may never have observed a particular \(M_iN_j\) combination. Nevertheless, it can learn that \(M_i\) has a distributional profile similar to those of other M tokens and that \(N_j\) occurs after those tokens. On this basis, the model can assign a high probability to \(N_j\) following \(M_i\), even though the specific combination is unattested. This explains the previous observations that the learning about one sequence can increase the probability of others (Bengio et al., 2003; Kapatsinski, 2026). Similar account could be seen in the statistical learning research in human cognitive science (Smith, 1969; Reeder et al., 2013, 2017).

Most importantly, this study demonstrates that categories cannot merely be defined by a set of discrete tokens. Instead, a category is a continuous region in the representation space. This geometric definition gives predictive significance to regions that contain no observed data points, allowing models to extend abstractions learned from observed inputs to unobserved combinations. This echoes with the findings of Marro et al. (2025) in LLMs, they show that empty regions between specific tokens are semantically continuous. For example, the empty regions closer to the token ``apple'' tend to share more semantic features of apple, and the empty regions closer to the token ``banana'' tend to share more semantic features of banana. In the current study, we extend the observations to category level representations.

Across the learning trajectory, the geometric structure associated with these categories emerges progressively from the earliest checkpoints. This supports the view that NLMs generalize from the beginning of learning, rather than doing so only after extensive memorization. Exemplar-specific representations play a minimal role until category-level representations are disrupted during over-fitting. This learning trajectory is observed in both Study 1 and Study 2. One possible interpretation is that updates to the representation space become progressively more localized: they initially affect the global representational landscape, later concentrate around observed data points, and eventually only cover the positions of those points. The global geometric structures in the representation space are therefore destructed when memorization becomes dominant.

This account does not contradict studies demonstrating that memorization constitutes part of the representation of NLMs (Carlini et al., 2021; Kandpal et al., 2022; Nasr et al., 2023). These studies demonstrate that NLMs are capable and do memorize input data points to a certain degree, especially when there is no generalizable patterns in the input. The current study instead directly examines the relation between memorization and generalization over the learning path. We simply demonstrate that when there are generalizable statistical patterns in the input, memorization does not dominate the early-stage learning and representation.

\section*{AI Use Statement}
Generative AI tools, including ChatGPT, were used for programming (including generating the synthetic grammar dataset and model training) and formatting (Latex compile).  All research questions, experimental designs, methodology, interpretation of results, and scientific claims were fully developed by the authors. All writing is conducted by the authors. All AI-assisted code and figures were independently checked by the authors, who take full responsibility for the final content of the paper.\par
\section*{References}
\begingroup\small
\noindent Arpit, D., Jastrz\k{e}bski, S., Ballas, N., Krueger, D., Bengio, E., Kanwal, M. S., Maharaj, T., Fischer, A., Courville, A., Bengio, Y., \& Lacoste-Julien, S. (2017). A closer look at memorization in deep networks. In \textit{Proceedings of the 34th International Conference on Machine Learning} (PMLR Vol. 70, pp. 233--242).\par\smallskip
\noindent Bayat, R., Pezeshki, M., Dohmatob, E., Lopez-Paz, D., \& Vincent, P. (2025). The pitfalls of memorization: When memorization hurts generalization. \textit{International Conference on Learning Representations}.\par\smallskip
\noindent Bengio, Y., Ducharme, R., Vincent, P., \& Jauvin, C. (2003). A neural probabilistic language model. \textit{Journal of Machine Learning Research, 3}, 1137--1155.\par\smallskip
\noindent Carlini, N., Liu, C., Erlingsson, \'{U}., Kos, J., \& Song, D. (2019). The secret sharer: Evaluating and testing unintended memorization in neural networks. In \textit{28th USENIX Security Symposium (USENIX Security 19)} (pp. 267--284). USENIX Association.\par\smallskip
\noindent Carlini, N., Tram\`{e}r, F., Wallace, E., Jagielski, M., Herbert-Voss, A., Lee, K., Roberts, A., Brown, T., Song, D., Erlingsson, \'{U}., Oprea, A., \& Raffel, C. (2021). Extracting training data from large language models. In \textit{30th USENIX Security Symposium (USENIX Security 21)} (pp. 2633--2650). USENIX Association.\par\smallskip
\noindent Carlini, N., Ippolito, D., Jagielski, M., Lee, K., Tram\`{e}r, F., \& Zhang, C. (2023). Quantifying memorization across neural language models. \textit{International Conference on Learning Representations}.\par\smallskip
\noindent Del\'{e}tang, G., Ruoss, A., Duquenne, P.-A., Catt, E., Genewein, T., Mattern, C., Grau-Moya, J., Wenliang, L. K., Aitchison, M., Orseau, L., Hutter, M., \& Veness, J. (2024). Language modeling is compression. \textit{International Conference on Learning Representations}.\par\smallskip
\noindent Feldman, V. (2020). Does learning require memorization? A short tale about a long tail. In \textit{Proceedings of the 52nd Annual ACM SIGACT Symposium on Theory of Computing} (pp. 954--959). Association for Computing Machinery. \url{https://doi.org/10.1145/3357713.3384290}\par\smallskip
\noindent Ghosh, B., Das, S., Wu, Q., Khan, M. A., Gummadi, K. P., Terzi, E., \& Garg, D. (2025). Rethinking memorization measures and their implications in large language models. \textit{arXiv preprint arXiv:2507.14777}.\par\smallskip
\noindent Hayes, J., Swanberg, M., Chaudhari, H., Yona, I., Shumailov, I., Nasr, M., Choquette-Choo, C. A., Lee, K., \& Cooper, A. F. (2025). Measuring memorization in language models via probabilistic extraction. In \textit{Proceedings of the 2025 Conference of the Nations of the Americas Chapter of the Association for Computational Linguistics: Human Language Technologies} (Vol. 1, pp. 9266--9291). Association for Computational Linguistics. \url{https://doi.org/10.18653/v1/2025.naacl-long.469}\par\smallskip
\noindent Jacot, A., Gabriel, F., \& Hongler, C. (2018). Neural tangent kernel: Convergence and generalization in neural networks. In \textit{Advances in Neural Information Processing Systems} (Vol. 31).\par\smallskip
\noindent Kandpal, N., Wallace, E., \& Raffel, C. (2022). Deduplicating training data mitigates privacy risks in language models. In \textit{Proceedings of the 39th International Conference on Machine Learning} (PMLR Vol. 162, pp. 10697--10707).\par\smallskip
\noindent Kapatsinski, V. (2026). Transformers perform adaptive partial pooling. \textit{arXiv preprint arXiv:2602.03980}.\par\smallskip
\noindent Kolmogorov, A. N. (1965). Three approaches to the quantitative definition of information. \textit{Problems of Information Transmission, 1}(1), 1--7.\par\smallskip
\noindent Marro, S., Evangelista, D., Huang, X. A., La Malfa, E., Lombardi, M., \& Wooldridge, M. J. (2025). Language models are implicitly continuous. \textit{International Conference on Learning Representations}.\par\smallskip
\noindent Morris, J. X., Sitawarin, C., Kokhlikyan, N., Guo, C., Suh, G. E., Rush, A. M., Chaudhuri, K., \& Mahloujifar, S. (2026). How much can language models memorize? In \textit{Proceedings of the 43rd International Conference on Machine Learning}.\par\smallskip
\noindent Nasr, M., Carlini, N., Hayase, J., Jagielski, M., Cooper, A. F., Ippolito, D., Choquette-Choo, C. A., Wallace, E., Tram\`{e}r, F., \& Lee, K. (2023). Scalable extraction of training data from (production) language models. \textit{arXiv preprint arXiv:2311.17035}.\par\smallskip
\noindent Novak, R., Bahri, Y., Abolafia, D. A., Pennington, J., \& Sohl-Dickstein, J. (2018). Sensitivity and generalization in neural networks: An empirical study. \textit{International Conference on Learning Representations}.\par\smallskip
\noindent Prashanth, U. S. S. N. S., Deng, A., O'Brien, K., S V, J., Khan, M. A., Borkar, J., Choquette-Choo, C. A., Fuehne, J. R., Biderman, S. R., Ke, T., Lee, K., \& Saphra, N. (2025). Recite, reconstruct, recollect: Memorization in LMs as a multifaceted phenomenon. \textit{International Conference on Learning Representations}.\par\smallskip
\noindent Reeder, P. A., Newport, E. L., \& Aslin, R. N. (2013). From shared contexts to syntactic categories: The role of distributional information in learning linguistic form-classes. \textit{Cognitive Psychology, 66}(1), 30--54. \url{https://doi.org/10.1016/j.cogpsych.2012.09.001}\par\smallskip
\noindent Reeder, P. A., Newport, E. L., \& Aslin, R. N. (2017). Distributional learning of subcategories in an artificial grammar: Category generalization and subcategory restrictions. \textit{Journal of Memory and Language, 97}, 17--29. \url{https://doi.org/10.1016/j.jml.2017.07.006}\par\smallskip
\noindent Schwarzschild, A., Feng, Z., Maini, P., Lipton, Z. C., \& Kolter, J. Z. (2024). Rethinking LLM memorization through the lens of adversarial compression. \textit{Advances in Neural Information Processing Systems, 37}.\par\smallskip
\noindent Shannon, C. E. (1948). A mathematical theory of communication. \textit{Bell System Technical Journal, 27}(3), 379--423; \textit{27}(4), 623--656. \url{https://doi.org/10.1002/j.1538-7305.1948.tb01338.x}\par\smallskip
\noindent Smith, K. H. (1969). Learning co-occurrence restrictions: Rule induction or rote learning? \textit{Journal of Verbal Learning and Verbal Behavior, 8}(2), 319--321. \url{https://doi.org/10.1016/S0022-5371(69)80086-1}\par\smallskip
\noindent Stephenson, C., Padhy, S., Ganesh, A., Hui, Y., Tang, H., \& Chung, S. (2021). On the geometry of generalization and memorization in deep neural networks. \textit{International Conference on Learning Representations}.\par\smallskip
\noindent T\"{a}nzer, M., Ruder, S., \& Rei, M. (2022). Memorisation versus generalisation in pre-trained language models. In \textit{Proceedings of the 60th Annual Meeting of the Association for Computational Linguistics} (Vol. 1, pp. 7564--7578). Association for Computational Linguistics. \url{https://doi.org/10.18653/v1/2022.acl-long.521}\par\smallskip
\noindent Zhang, C., Bengio, S., Hardt, M., Recht, B., \& Vinyals, O. (2017). Understanding deep learning requires rethinking generalization. \textit{International Conference on Learning Representations}.\par\smallskip
\noindent Zhang, C., Ippolito, D., Lee, K., Jagielski, M., Tram\`{e}r, F., \& Carlini, N. (2023). Counterfactual memorization in neural language models. \textit{Advances in Neural Information Processing Systems, 36}, 39321--39362.\par\smallskip
\endgroup
\clearpage
\appendix
\section*{Materials and Methods}
\subsection*{Study 1}
\subsubsection*{Synthetic language}
The Study 1 vocabulary comprised 250 tokens divided into five mutually exclusive categories, M, N, P, Q, and U, with 50 tokens in each category. The grammar contained two local dependencies: every M token could be followed by every N token, and every P token could be followed by every Q token. It therefore licensed 2,500 distinct MN bigrams and 2,500 distinct PQ bigrams.\par
We generated 10,000 five-token sequences by first sampling every position independently and uniformly from the U category. We then selected 5,000 sequences without replacement. In each selected sequence, one of the four adjacent position pairs was chosen and the two U tokens were replaced with one MN or PQ bigram. Each of the 5,000 grammatical bigrams was used exactly once; the remaining 5,000 sequences contained only U tokens.\par
After generation, the completed sequences were randomly permuted and divided into 8,000 training sequences, 1,000 validation sequences, and 1,000 test sequences. Because each MN or PQ type occurred only once before the split, structured test sequences contained bigrams absent from training. Their component tokens nevertheless occurred in other training bigrams, allowing category-level generalization to be evaluated without withholding the individual lexical items.\par
\subsubsection*{Models and training}
We trained three causal decoder-only Transformers: 2 layers, 2 attention heads, and an 8-dimensional embedding space (2-2-8); 4 layers, 4 heads, and a 32-dimensional embedding space (4-4-32); and 4 layers, 4 heads, and a 128-dimensional embedding space (4-4-128). Each model used learned token and positional embeddings, pre-layer-normalized causal self-attention, a GELU feed-forward network with width four times the embedding dimension, and a final layer normalization followed by a full-vocabulary output projection. Input and output embeddings were not tied, and dropout and weight decay were set to zero.\par
Models were optimized with next-token cross-entropy using Adam, with a learning rate of 0.001, \ensuremath{\beta}1 = 0.9, \ensuremath{\beta}2 = 0.95, \ensuremath{\varepsilon} = 10\textsuperscript{-8}, and a batch size of 64. The 2-2-8 analysis contained 101 saved checkpoints, including initialization. The 4-4-32 and 4-4-128 analyses each contained 591 saved checkpoints, including initialization.\par
\subsubsection*{Checkpoint-specific geometric analysis}
At each saved checkpoint, we extracted the static input embeddings and independently calculated the geometric centres of the M and P token embeddings:\par
\[M_c = \frac{1}{\lvert M\rvert}\sum_{m\in M}e_m,\]
\[P_c = \frac{1}{\lvert P\rvert}\sum_{p\in P}e_p.\]
At each checkpoint, we connected the two centres with the interpolation line:\par
\[x(t) = (1-t)M_c+tP_c,\qquad 0\leq t\leq 1.\]
We sampled 200 equally spaced values of t. The resulting interpolation vectors were not vocabulary items and therefore probed continuous regions of embedding space containing no observed token. We sampled 32 five-token U-only backgrounds and inserted each vector into each of the four non-final positions, yielding 128 contexts for every value of t. We calculated the next-token probability mass assigned to all N tokens and, separately, all Q tokens, then averaged each mass over the 128 contexts. The learning-path heatmaps report log P(N) \ensuremath{-} log P(Q). The M and P centres were recomputed independently at every checkpoint.\par
\subsubsection*{Observed-token analysis}
We also measured category prediction for observed lexical tokens. For every held-out test sequence containing an MN bigram, we evaluated the next-token distribution after the M token and summed the probabilities assigned to all N tokens. We applied the analogous procedure to PQ test sequences and the Q category. Means were calculated separately for the two relations at every saved checkpoint; 0.2 is the uniform five-category reference.\par
Each Study 1 configuration was evaluated in three independent replicates. The main text reports the first replicate for each model, and the remaining analyses will be presented in the Appendix.\par
\subsection*{Study 2}
\subsubsection*{Synthetic language}
The ABCDE core has 10\textsuperscript{5} = 100,000 possible lexical combinations. For each replicate, we sampled disjoint sets of 50,000 training cores, 5,000 validation cores, and 5,000 test cores. Each selected core was combined with all 10 U tokens in both permitted positions, so every core produced 20 surface strings. The resulting dataset contained 1,000,000 training strings, 100,000 validation strings, and 100,000 test strings. Held-out cores therefore contained ABCDE combinations absent from training, while their component tokens were represented in the training data.\par
\subsubsection*{Models and training}
We trained three causal decoder-only Transformers: 1 layer, 1 attention head, and an 8-dimensional representation (1-1-8); 2 layers, 2 heads, and a 32-dimensional representation (2-2-32); and 4 layers, 4 heads, and a 64-dimensional representation (4-4-64). The models used learned token and positional embeddings, pre-layer-normalized causal self-attention, GELU feed-forward networks with width four times the model dimension. Dropout and weight decay were zero.\par
Models were trained using Adam, with a learning rate of 0.0003, \ensuremath{\beta}1 = 0.9, \ensuremath{\beta}2 = 0.999, \ensuremath{\varepsilon} = 10\textsuperscript{-8}, and a batch size of 128. No learning-rate schedule or gradient clipping was used. The 1,000,000 training strings were arranged in one random permutation and presented cyclically in contiguous minibatches. For the analyses reported here, all models were compared through iteration 1,000,000. Information-compression metrics were evaluated every 100 iterations.\par
\subsubsection*{Information-compression analysis}
Following the information-compression perspective of Morris et al. (2026), we quantify model knowledge in bits. In information theory, the Shannon self-information of a sequence is the negative base-2 logarithm of its probability (Shannon, 1948). For an autoregressive language model, this quantity is the sum of the token-level surprisals and corresponds to the ideal code length assigned by a compressor based on the model's probabilities. Morris et al. motivate such likelihood-based code lengths as practical approximations to conditional Kolmogorov complexity---the length of the shortest description of a sequence when a model is available---because exact Kolmogorov complexity is uncomputable (Kolmogorov, 1965). We adapt this perspective rather than implement their complete unintended-memorization estimator: we compare the information assigned by the same model to familiar training strings and held-out grammatical strings, using the category-level grammar as a theoretical reference.\par
For a six-token string x = (x\textsubscript{1}, \ldots{}, x\textsubscript{6}) at checkpoint s, model \ensuremath{\theta}\textsubscript{s} assigns a probability to each token conditional on the preceding tokens. We defined the information content of the complete sequence as:\par
\[I_s(x)=-\sum_{t=1}^{6}\log_2 p_{\theta_s}(x_t\mid x_{<t})\]
We define the information-compression score as\par
\[C_s(x)=I_s(x)-I_{\mathrm{naive}}(x)\]
such that the na\"{i}ve baseline is 0 bits and lower, more negative values indicate stronger compression. Lower information content therefore corresponds to higher model probability and stronger compression. Because the measure is expressed in bits, differences in information can be interpreted as differences in the length of the code assigned to a sequence by the model.\par
To trace changes in information compression across training, we constructed two fixed probe sets within each replicate. The familiar probe set contained 4,096 distinct ABCDE cores sampled from the 50,000 training cores. The generalization probe set contained 4,096 distinct cores sampled from the held-out test set. These held-out cores were combinations of A, B, C, D, and E tokens that never occurred in training, although all their individual component tokens appeared in other training sequences. The distinction between the two probe sets therefore concerned whether a particular lexical combination had been observed, rather than whether its component tokens were known to the model.\par
The same probe cores were evaluated at every checkpoint, ensuring that changes in information reflected changes in the model rather than changes in the evaluation sample. Each core was combined with every one of the 10 U tokens in both licensed positions: U could occur either before the ABCDE core or after it. Each core therefore generated 10 \ensuremath{\times} 2 = 20 grammatical surface strings. We calculated the information of every surface string, averaged across the 20 surface forms associated with each core, and then averaged across the 4,096 cores. This procedure gave every core equal weight and prevented the trajectory from being driven by a particular U token or U position. The trajectory calculated from training cores is labelled familiar, whereas the trajectory calculated from held-out cores is labelled generalization.\par
We used two theoretical baselines corresponding to two idealized states of model knowledge: a completely naive model and a model with perfect knowledge of the underlying data-generating process.\par
A completely naive model has learned neither the grammatical orders nor the token categories. It assigns equal probability to every token in the full 60-token vocabulary at each of the six positions. The probability it assigns to any six-token sequence is therefore:\par
\[P_{\mathrm{naive}}(x)=\left(\frac{1}{60}\right)^6\]
The corresponding information content is:\par
\[I_{\mathrm{naive}}(x)=6\log_2(60).\]
This gives 6 log\textsubscript{2} 60 = 35.44 bits. This naive-model baseline represents the information expected before the model has learned any systematic property of the artificial language.\par
A model with perfect generalization knows the true category-level data-generating process. It knows that U can occur at either boundary, that the remaining positions follow the ABCDE category order, and that each required category contains 10 equally probable tokens. However, it has no knowledge of which particular ABCDE cores occurred in the training sample. Such a model assigns equal probability to the two grammatical orders and equal probability to each of the k = 10 tokens in the six required category positions. The probability of any grammatical sequence is therefore:\par
\[P_{\mathrm{grammar}}(x)=\frac{1}{2k^6}.\]
The corresponding perfect-generalization information content is:\par
\[I_{\mathrm{grammar}}(x)=\log_2(2k^6)=1+6\log_2 k.\]
For k = 10, this gives log\textsubscript{2}(2 \ensuremath{\times} 10\textsuperscript{6}) = 20.93 bits. The additional one bit represents the choice between placing U at the beginning or at the end. Because the true data-generating process is uniform across the two orders and the tokens within each category, 20.93 bits is the strongest compression that can be justified by perfect category-level generalization alone.\par
The learning trajectories can therefore be interpreted relative to these two baselines. A decrease from the 0-bit naive-model baseline toward the -14.51-bit perfect-generalization baseline indicates that the model is learning the grammatical categories and their ordering. If familiar and held-out sequences decrease together, this learning generalizes beyond the particular combinations encountered during training. Systematic compression of familiar strings below the -14.51-bit baseline cannot be explained by improved knowledge of the underlying grammar, because that baseline already represents complete knowledge of the true data-generating process. It instead indicates that the model has acquired additional information about which particular combinations occurred in the training data. We therefore attribute this additional familiar-string compression to exemplar-specific memorization. A growing separation in which familiar sequence information-compression score falls below the perfect-generalization baseline while held-out sequence information-compression score remains higher provides correspondingly stronger evidence that memorization is becoming dominant and generalization is being compromised.\par
\subsubsection*{Sampled ranking analysis}
For the sampled ranking analysis, we used 200 saved checkpoints from iteration 5,000 through iteration 1,000,000. At each checkpoint, the observed pool comprised unique training strings whose first exposure occurred by that iteration. We independently sampled with replacement 1,024 strings from this pool and 512 strings from the held-out test set. Both samples were redrawn at every checkpoint using deterministic seed-, model-, checkpoint-, and pool-specific random-number streams. For each held-out string, we calculated the percentage of sampled observed strings with strictly lower information and then averaged this percentage over the 512 held-out draws. A value of 50\% is the equal-distribution reference.\par
For each sampled held-out string u, the checkpoint-specific percentage can be written as:\par
\[R_s(u)=100\times\frac{\#\{o:I_s(o)<I_s(u)\}}{1{,}024}.\]
All Study 2 analyses were completed for three independent replicates in which the data split, training order, model initialization, and evaluation-probe seeds varied. The main text reports the first replicate; the remaining analyses will be presented in the Appendix.\par
\clearpage
\section*{Appendix}
This Appendix presents the complete three-seed analyses for both studies. Within each figure set, the in-text replicate (Seed 1) is shown first, followed by Seeds 2 and 3. A single overall caption follows each three-seed set.\par
\subsection*{Study 1 multi-seed analyses}
\subsubsection*{2-2-8 model}
All three 2-2-8 runs form a smooth category boundary within the 5,000-iteration analysis horizon. In parallel, the correct target-category probability mass for observed MN and PQ prompts rises well above the 0.2 baseline in every run. The rapid organization reported in the main text is therefore reproduced across seeds.\par
\noindent\textbf{Seed 1: }\href{https://study1-trajectories-nx47.bojun-fl.chatgpt.site/model-2-2-8.html}{\textcolor{blue}{\underline{click to view animated visualization}}}\par
\begin{center}\includegraphics[width=\textwidth,height=0.70\textheight,keepaspectratio]{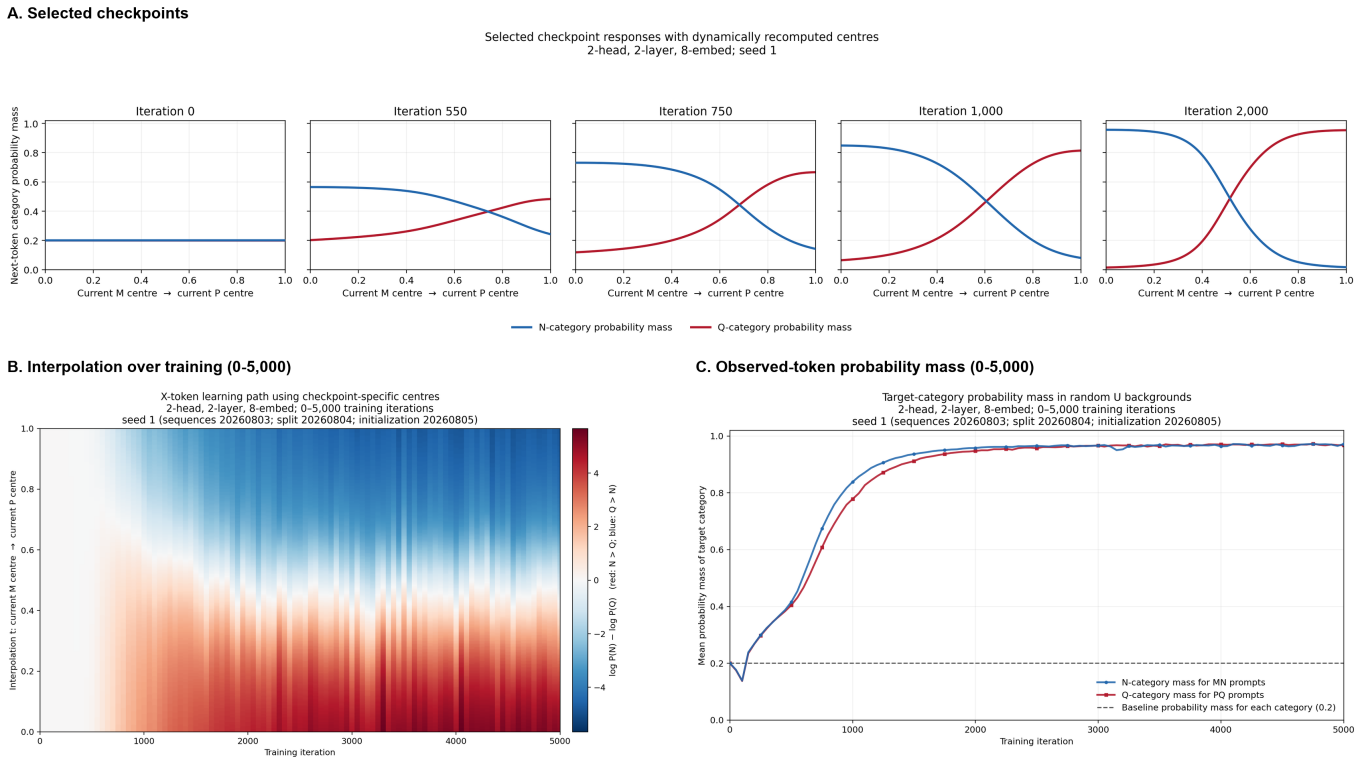}\end{center}
\clearpage
\noindent\textbf{Seed 2: }\href{https://study1-trajectories-nx47.bojun-fl.chatgpt.site/model-2-2-8-seed-2.html}{\textcolor{blue}{\underline{click to view animated visualization}}}\par
\begin{center}\includegraphics[width=\textwidth,height=0.70\textheight,keepaspectratio]{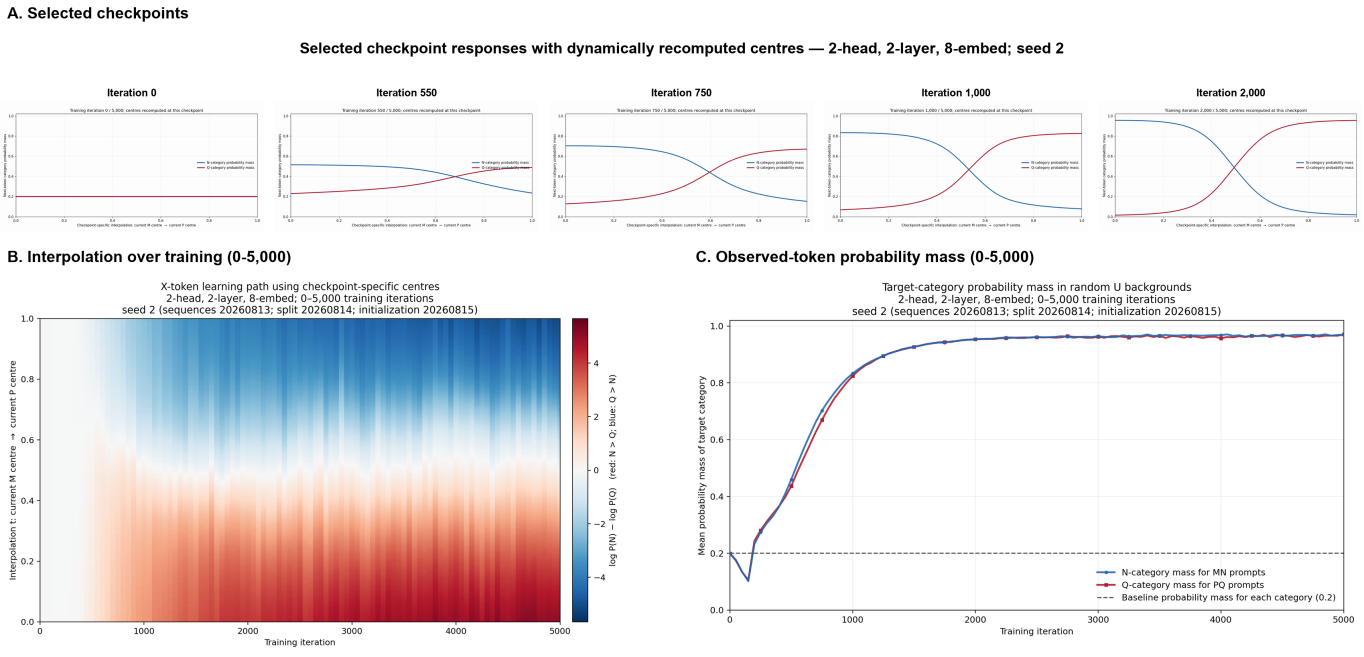}\end{center}
\clearpage
\noindent\textbf{Seed 3: }\href{https://study1-trajectories-nx47.bojun-fl.chatgpt.site/model-2-2-8-seed-3.html}{\textcolor{blue}{\underline{click to view animated visualization}}}\par
\begin{center}\includegraphics[width=\textwidth,height=0.70\textheight,keepaspectratio]{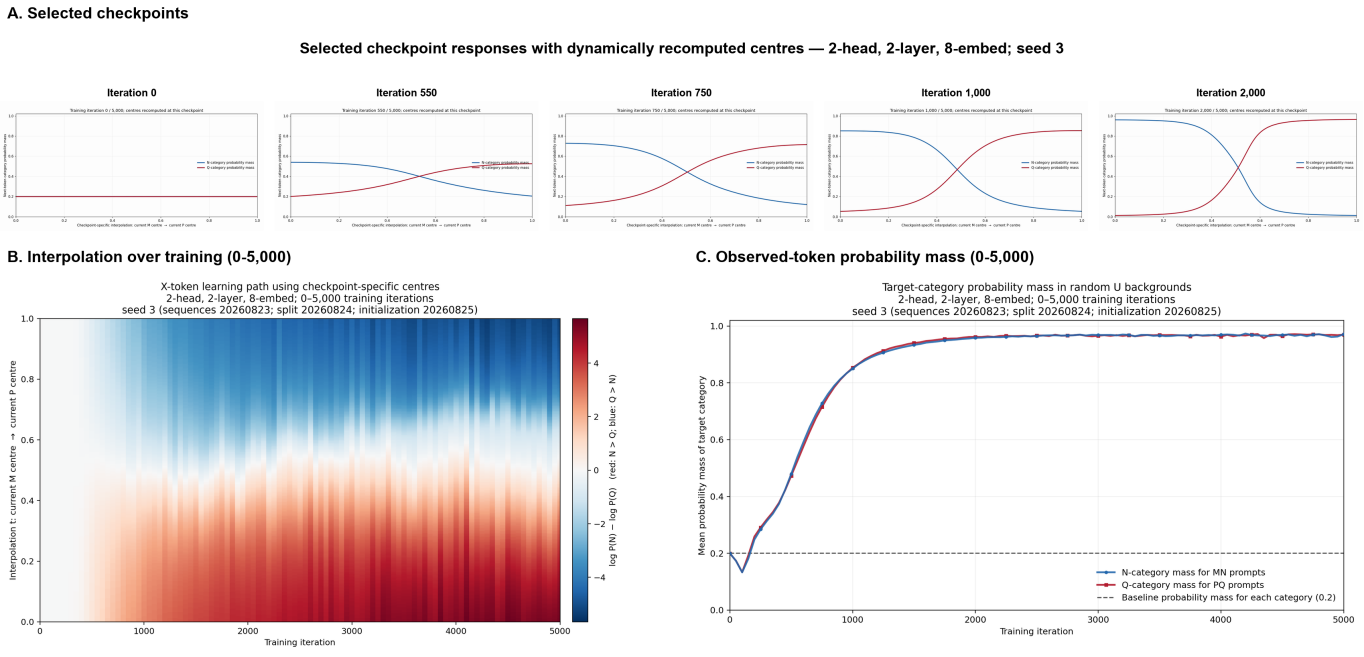}\end{center}
\noindent Appendix Figure A1. Multi-seed results for the 2-2-8 model. Complete results for Seeds 1, 2, and 3 are shown in that order. Each seed includes (A) selected interpolation checkpoints at iterations 0, 550, 750, 1,000, and 2,000; (B) the dynamic-centre heatmap from 0 to 5,000 iterations; and (C) correct target-category probability mass for observed MN and PQ prompts.\par
\clearpage
\subsubsection*{4-4-32 model}
Across all three seeds, the 4-4-32 model develops a broad interpolation boundary that persists over the 500,000-iteration analysis horizon, while observed MN and PQ prompts continue to receive strong correct-category probability mass. The precise boundary position varies, but the qualitative pattern reported in the main text is stable.\par
\noindent\textbf{Seed 1: }\href{https://study1-trajectories-nx47.bojun-fl.chatgpt.site/model-4-4-32.html}{\textcolor{blue}{\underline{click to view animated visualization}}}\par
\begin{center}\includegraphics[width=\textwidth,height=0.70\textheight,keepaspectratio]{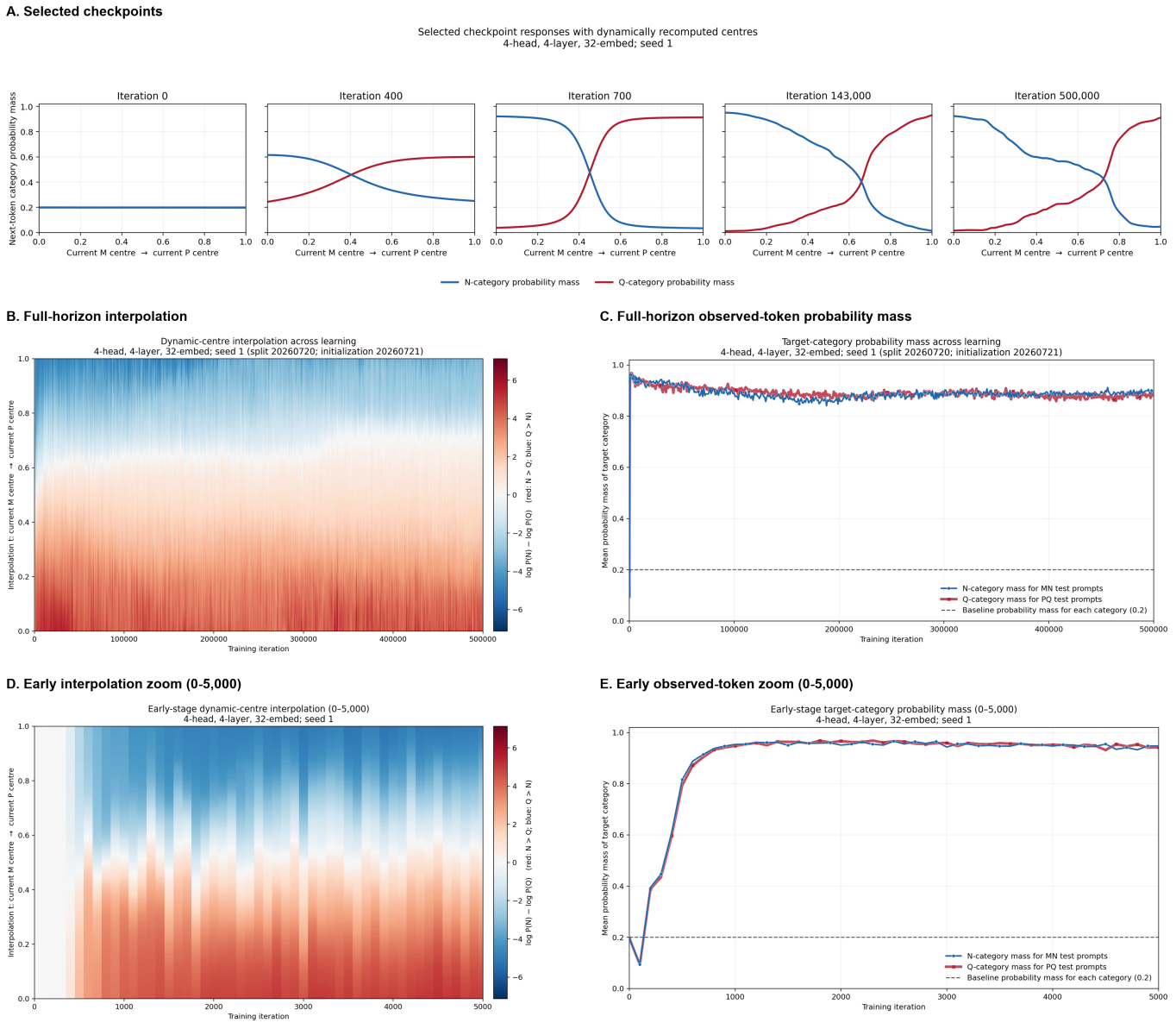}\end{center}
\clearpage
\noindent\textbf{Seed 2: }\href{https://study1-trajectories-nx47.bojun-fl.chatgpt.site/model-4-4-32-seed-2.html}{\textcolor{blue}{\underline{click to view animated visualization}}}\par
\begin{center}\includegraphics[width=\textwidth,height=0.70\textheight,keepaspectratio]{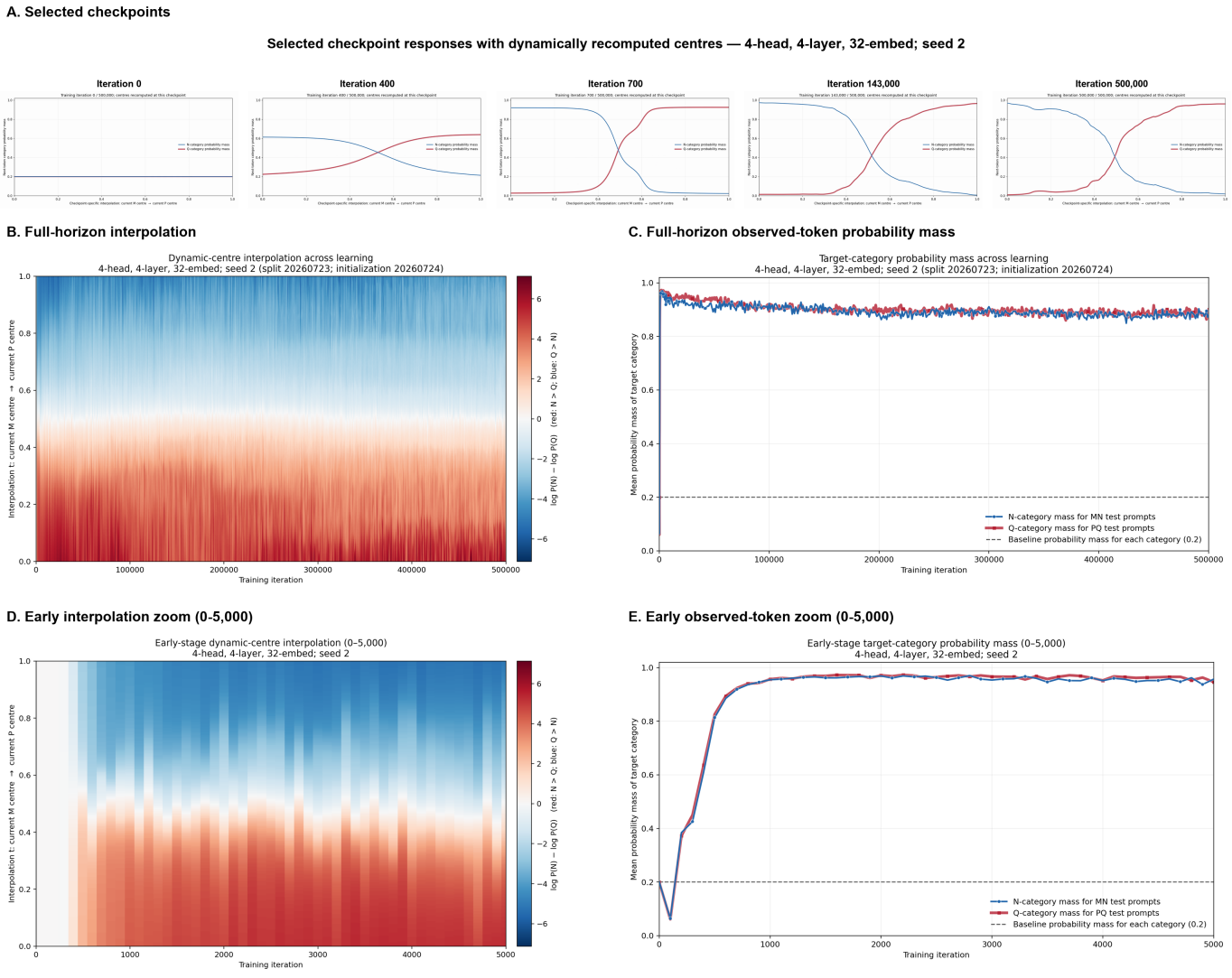}\end{center}
\clearpage
\noindent\textbf{Seed 3: }\href{https://study1-trajectories-nx47.bojun-fl.chatgpt.site/model-4-4-32-seed-3.html}{\textcolor{blue}{\underline{click to view animated visualization}}}\par
\begin{center}\includegraphics[width=\textwidth,height=0.70\textheight,keepaspectratio]{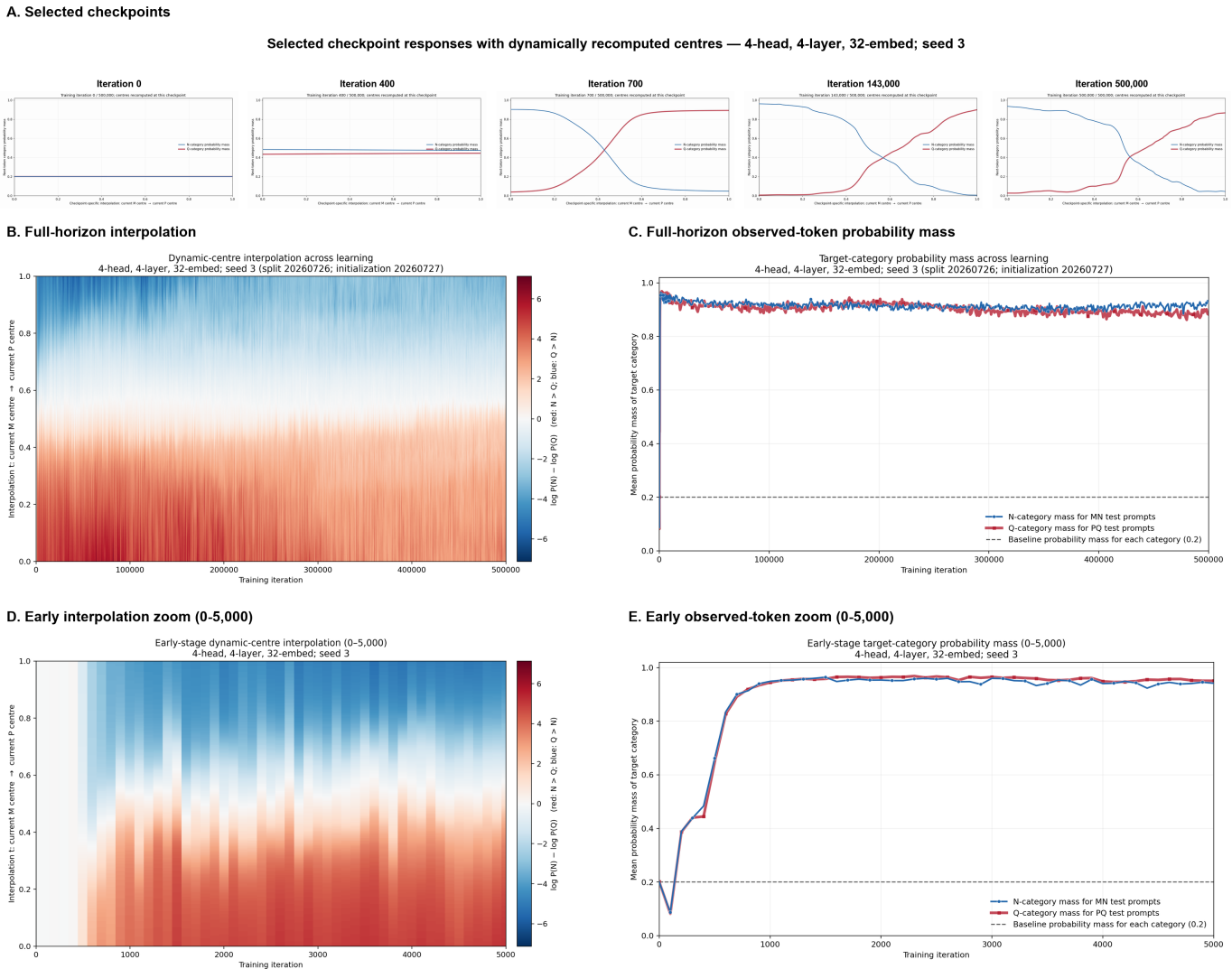}\end{center}
\noindent Appendix Figure A2. Multi-seed results for the 4-4-32 model. Complete results for Seeds 1, 2, and 3 are shown in that order. Each seed includes (A) selected interpolation checkpoints at iterations 0, 400, 700, 143,000, and 500,000; (B) the full dynamic-centre heatmap; (C) correct target-category probability mass for observed MN and PQ prompts; (D) the 0--5,000 interpolation close-up; and (E) the corresponding observed-prompt close-up.\par
\clearpage
\subsubsection*{4-4-128 model}
The three 4-4-128 runs consistently show early organization followed by a weaker and more fragmented response in the interpolated representation space during prolonged training. Responses to observed MN and PQ prompts remain above the 0.2 baseline. The dissociation between weakening global geometry and retained observed-token learning is therefore present across seeds.\par
\noindent\textbf{Seed 1: }\href{https://study1-trajectories-nx47.bojun-fl.chatgpt.site/model-4-4-128.html}{\textcolor{blue}{\underline{click to view animated visualization}}}\par
\begin{center}\includegraphics[width=\textwidth,height=0.70\textheight,keepaspectratio]{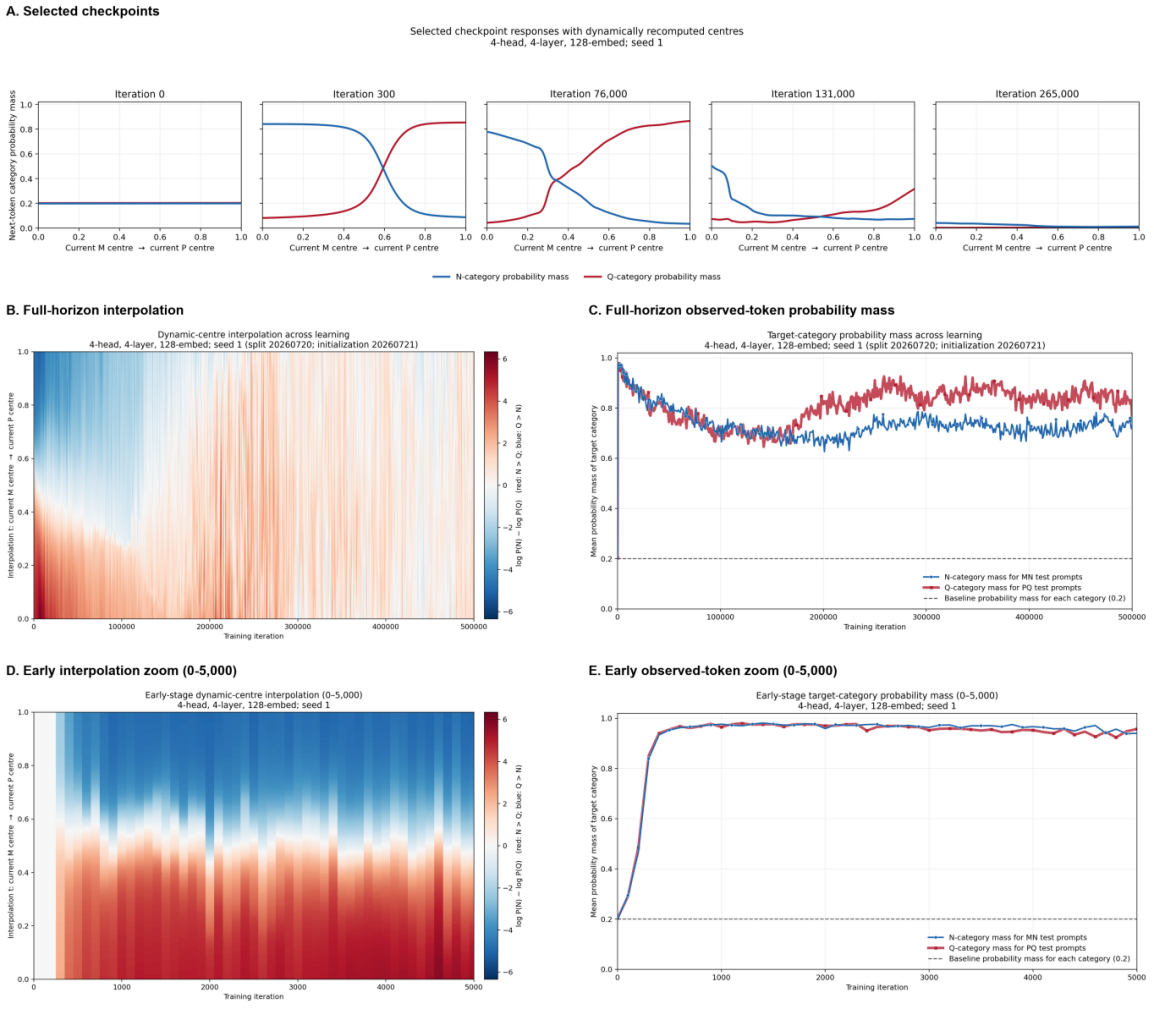}\end{center}
\clearpage
\noindent\textbf{Seed 2: }\href{https://study1-trajectories-nx47.bojun-fl.chatgpt.site/model-4-4-128-seed-2.html}{\textcolor{blue}{\underline{click to view animated visualization}}}\par
\begin{center}\includegraphics[width=\textwidth,height=0.70\textheight,keepaspectratio]{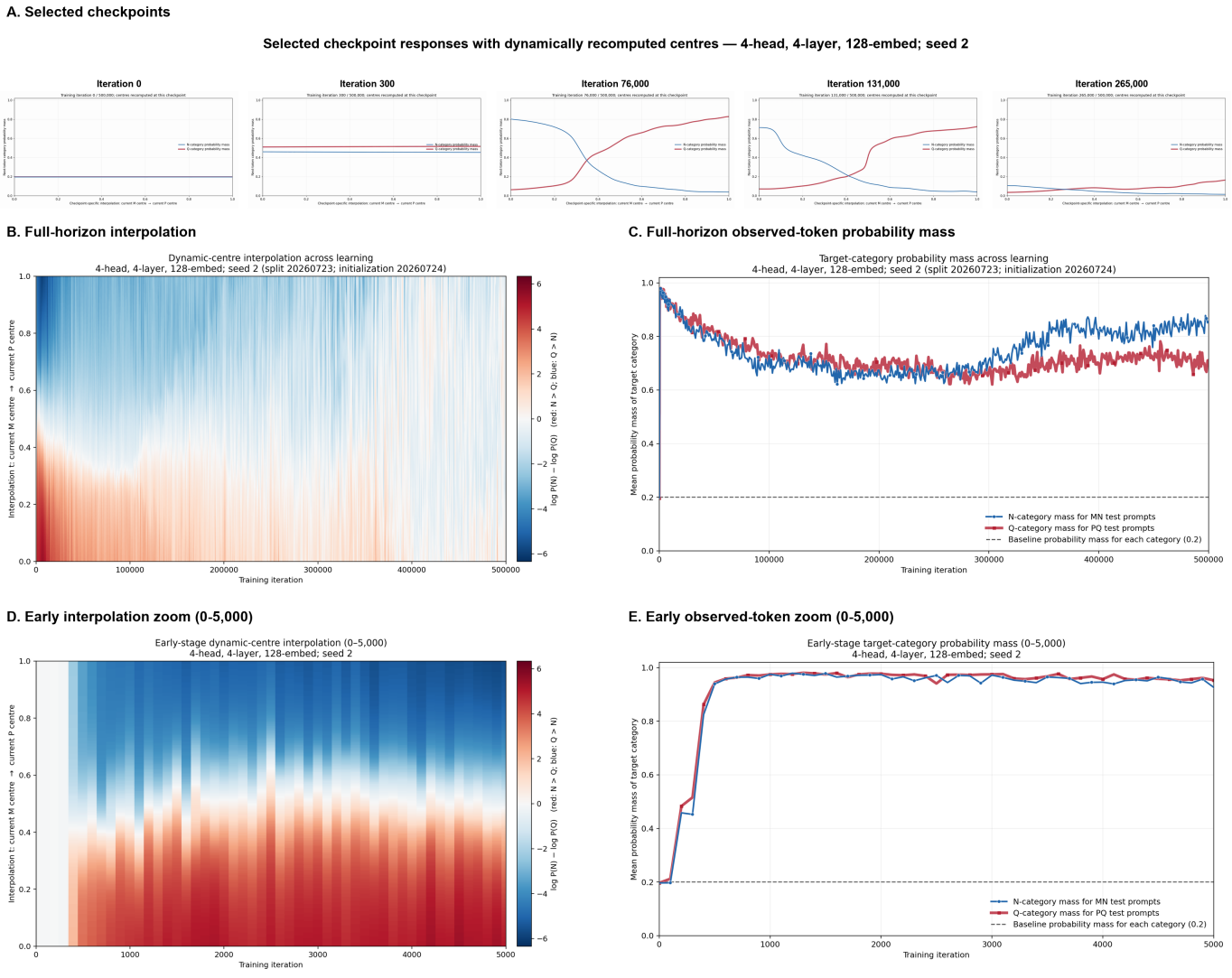}\end{center}
\clearpage
\noindent\textbf{Seed 3: }\href{https://study1-trajectories-nx47.bojun-fl.chatgpt.site/model-4-4-128-seed-3.html}{\textcolor{blue}{\underline{click to view animated visualization}}}\par
\begin{center}\includegraphics[width=\textwidth,height=0.70\textheight,keepaspectratio]{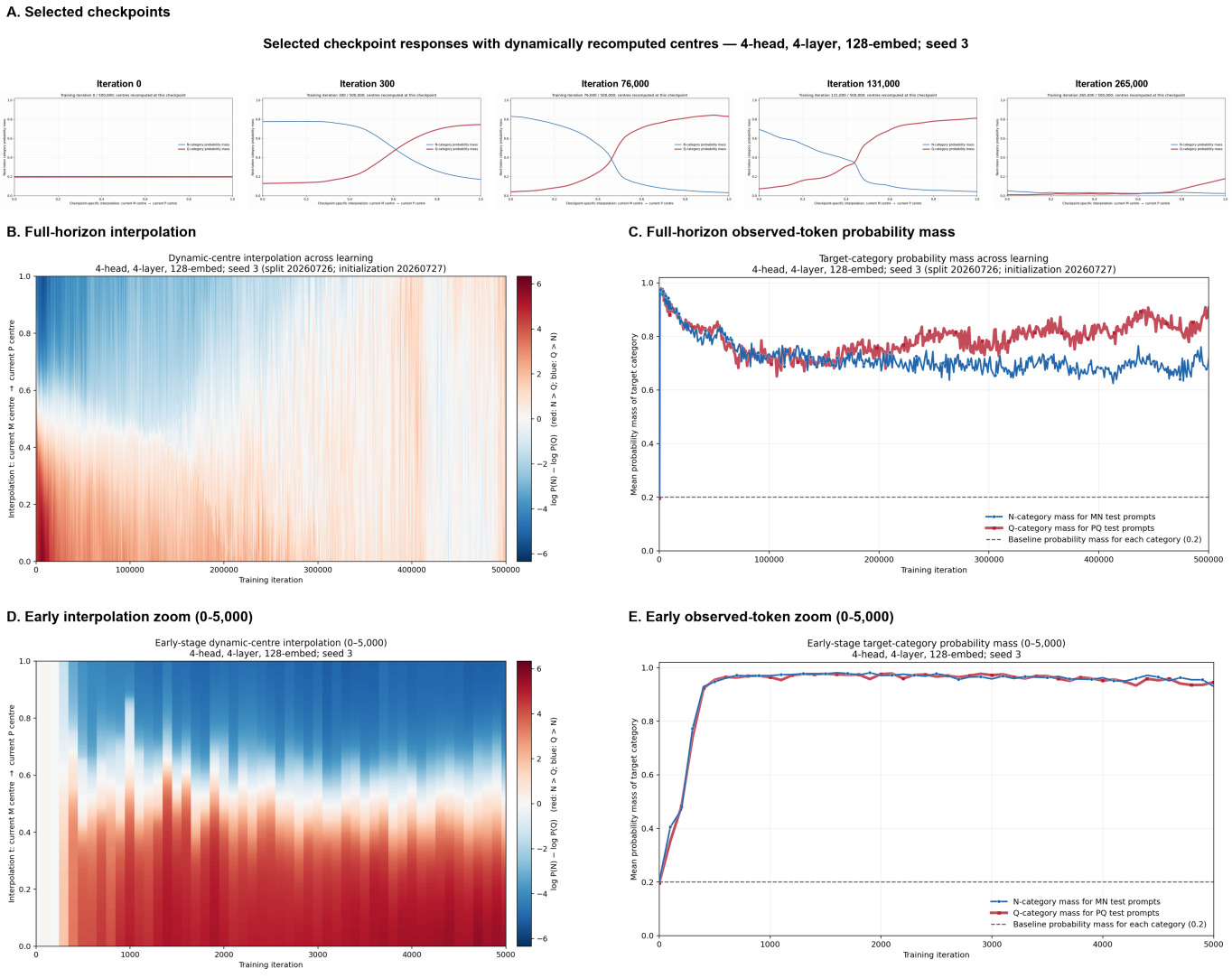}\end{center}
\noindent Appendix Figure A3. Multi-seed results for the 4-4-128 model. Complete results for Seeds 1, 2, and 3 are shown in that order. Each seed includes (A) selected interpolation checkpoints at iterations 0, 300, 76,000, 131,000, and 265,000; (B) the full dynamic-centre heatmap; (C) correct target-category probability mass for observed MN and PQ prompts; (D) the 0--5,000 interpolation close-up; and (E) the corresponding observed-prompt close-up.\par
\clearpage
\subsection*{Study 2 multi-seed analyses}
The six-token Study 2 pattern is reproduced across the three independent replicates. In every seed, the 1-1-8 model keeps familiar and held-out information-compression scores near the -14.51-bit category-level reference, the 2-2-32 model shows a moderate familiar--held-out divergence, and the 4-4-64 model shows the strongest divergence.\par
\subsubsection*{Seed 1}
\noindent\textbf{Information compression}\par
\begin{center}\includegraphics[width=\textwidth,height=0.70\textheight,keepaspectratio]{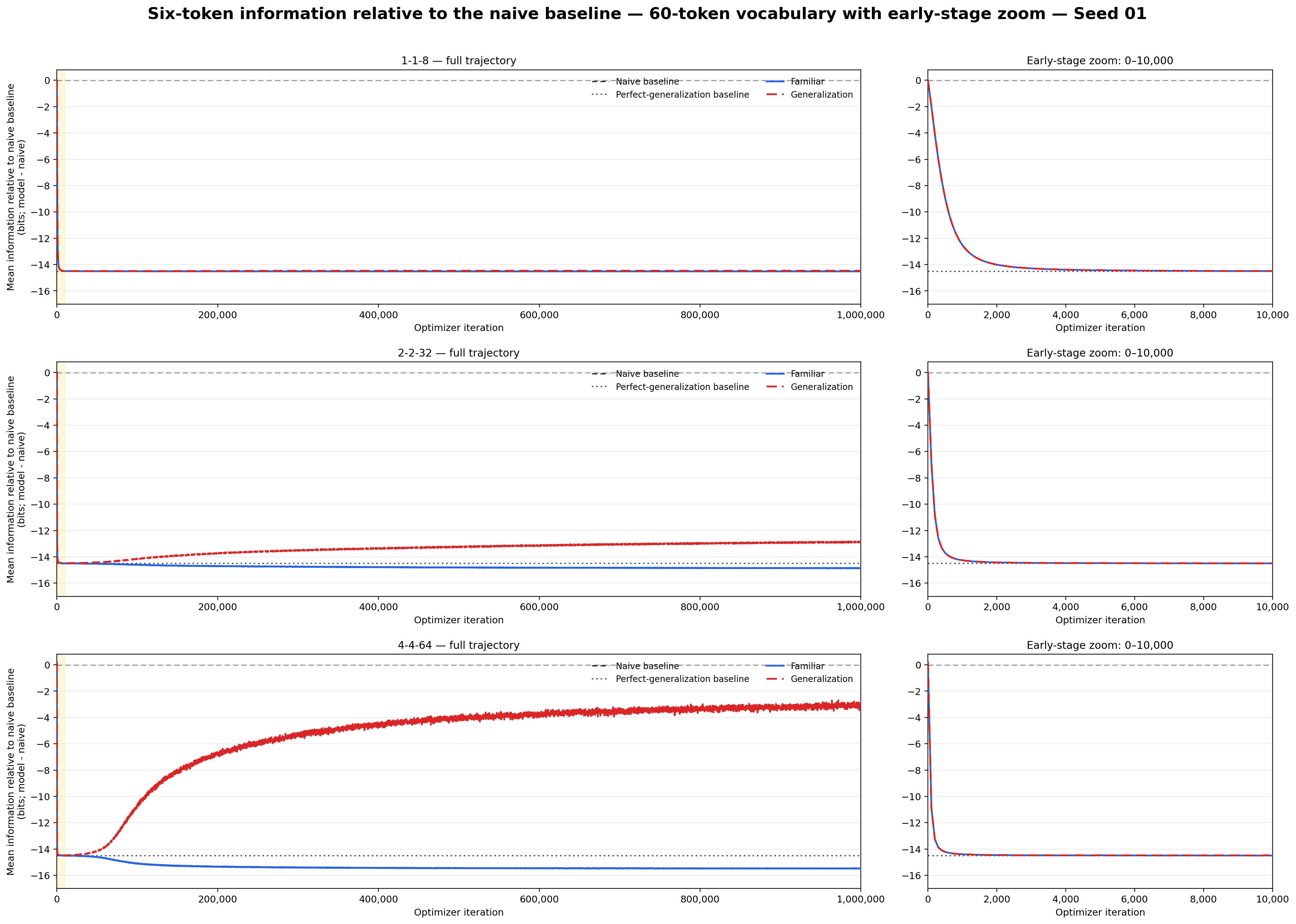}\end{center}
\clearpage
\noindent\textbf{Sampled ranking}\par
\begin{center}\includegraphics[width=\textwidth,height=0.70\textheight,keepaspectratio]{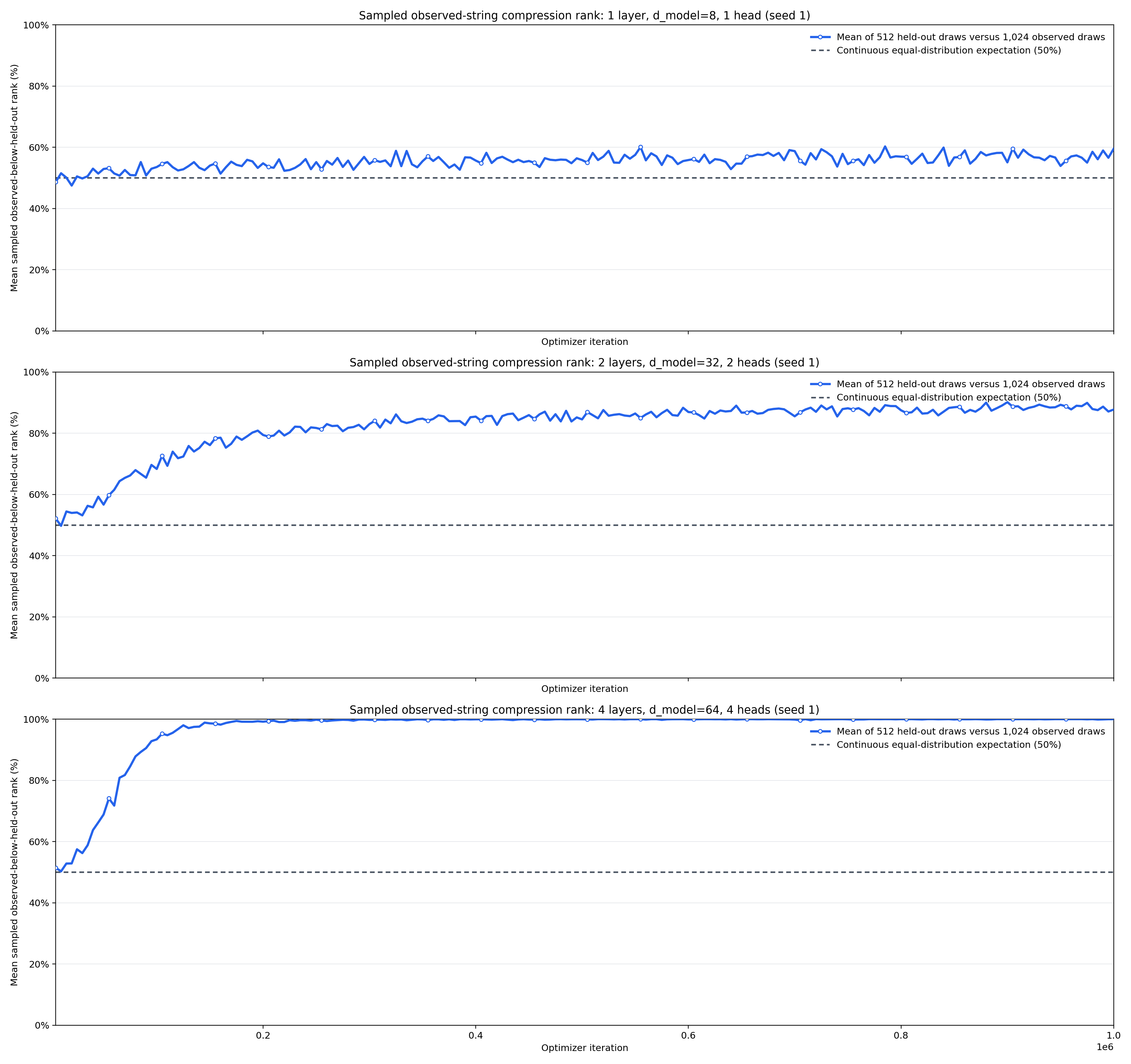}\end{center}
\clearpage
\subsubsection*{Seed 2}
\noindent\textbf{Information compression}\par
\begin{center}\includegraphics[width=\textwidth,height=0.70\textheight,keepaspectratio]{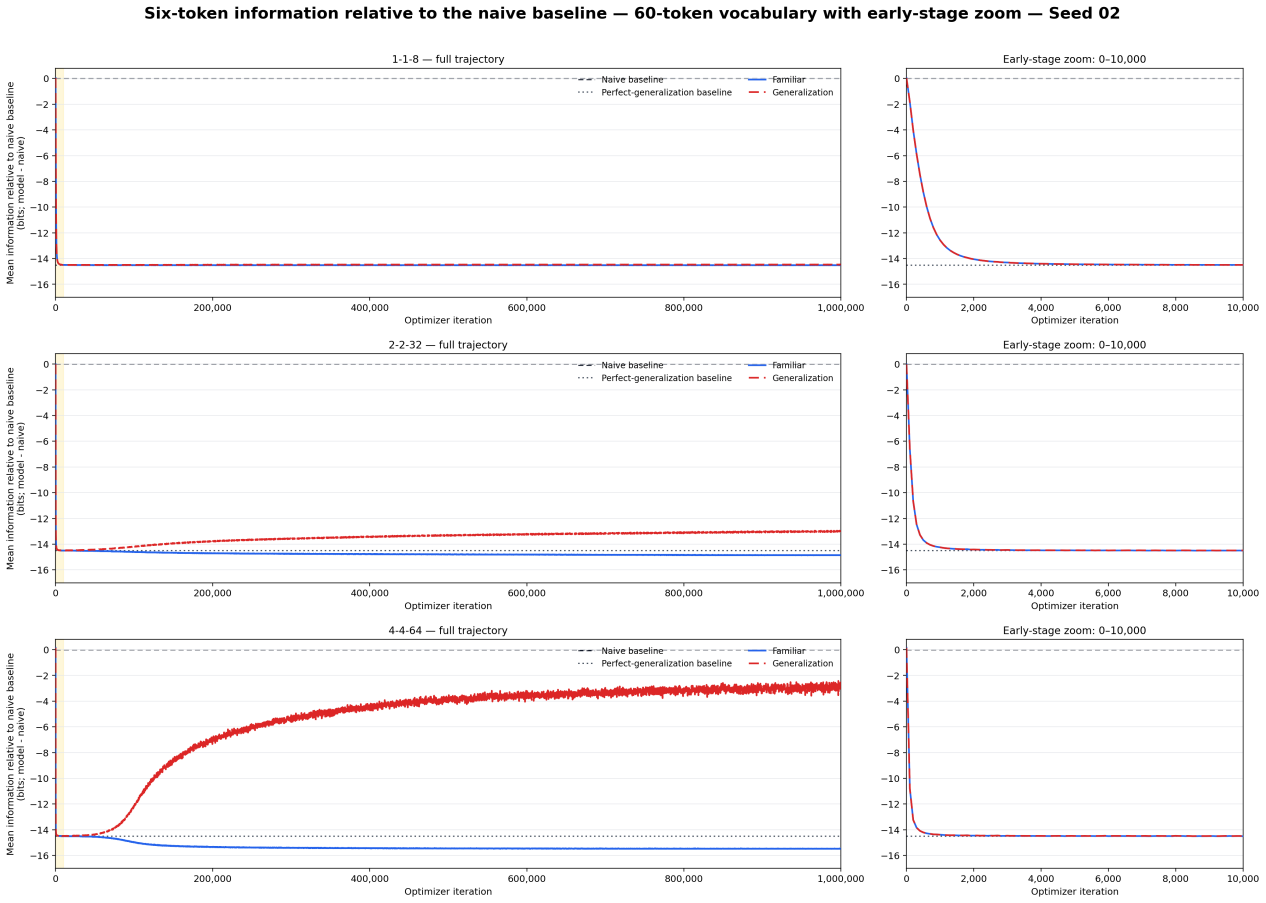}\end{center}
\clearpage
\noindent\textbf{Sampled ranking}\par
\begin{center}\includegraphics[width=\textwidth,height=0.70\textheight,keepaspectratio]{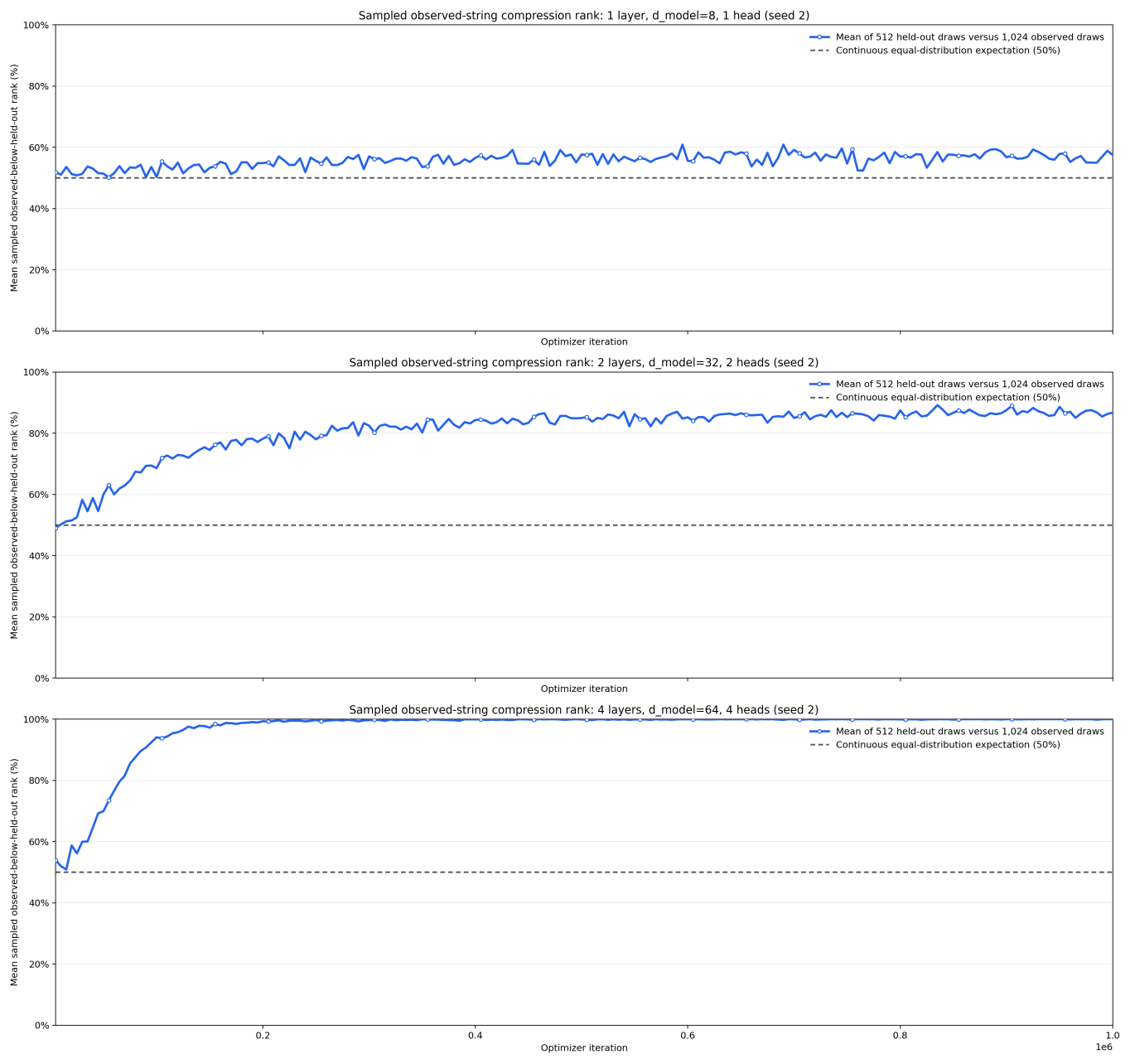}\end{center}
\clearpage
\subsubsection*{Seed 3}
\noindent\textbf{Information compression}\par
\begin{center}\includegraphics[width=\textwidth,height=0.70\textheight,keepaspectratio]{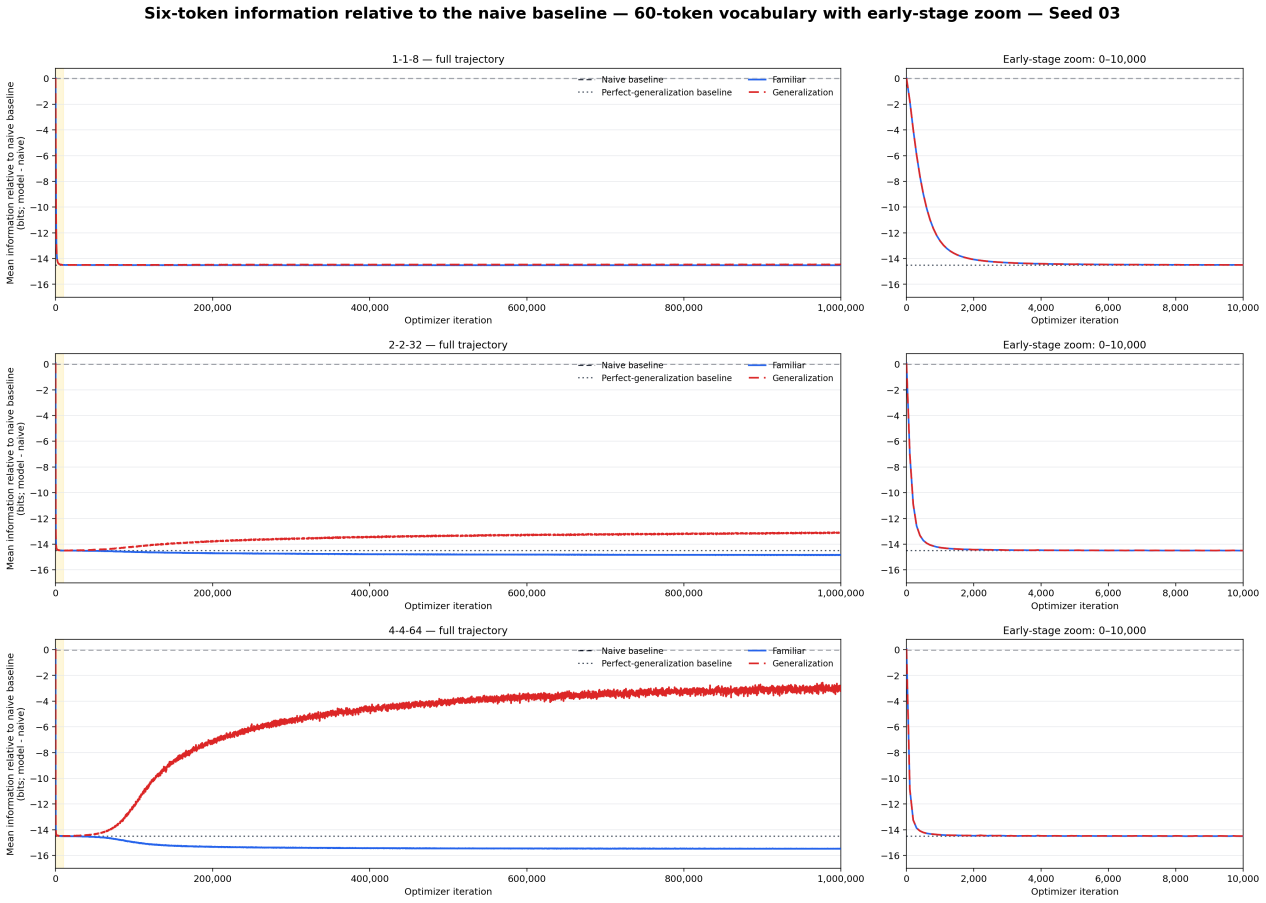}\end{center}
\clearpage
\noindent\textbf{Sampled ranking}\par
\begin{center}\includegraphics[width=\textwidth,height=0.70\textheight,keepaspectratio]{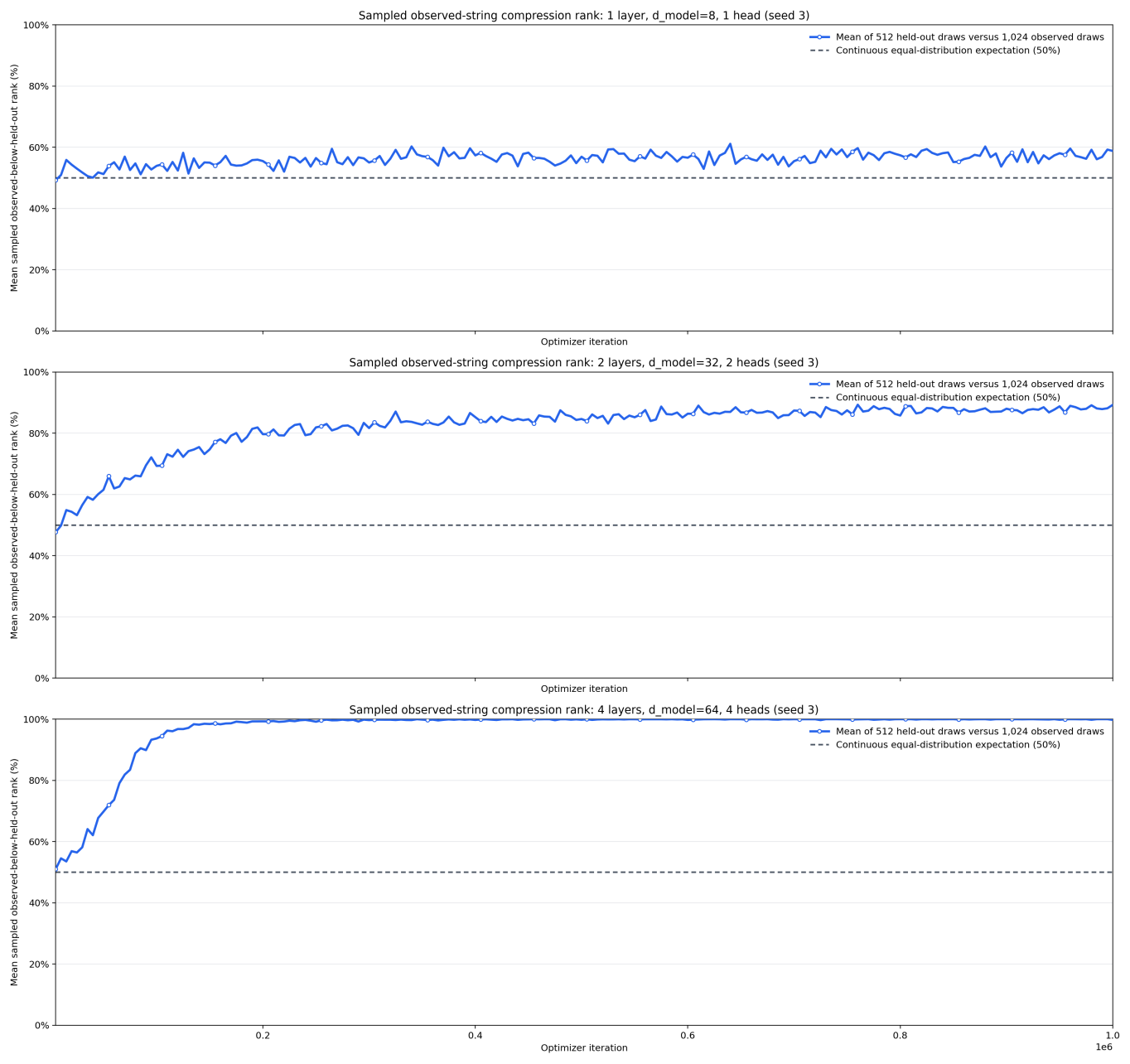}\end{center}
\noindent Appendix Figure A4. Multi-seed results for the six-token UABCDE/ABCDEU experiment. Complete results for Seeds 1, 2, and 3 are shown in that order. For each seed, the information-compression figure is followed by the sampled observed-string ranking figure. The information-compression panels compare familiar training-core strings with held-out test strings and mark the -14.51-bit category-level reference on the baseline-relative scale. The ranking panels show the mean percentage of sampled observed strings more strongly compressed than each held-out test string and mark the 50\% equal-distribution reference.\par

\end{document}